\documentclass[twoside,leqno,twocolumn]{article}

\usepackage[letterpaper]{geometry}

\usepackage{ltexpprt}
\usepackage{hyperref}

\usepackage{algorithm}
\usepackage{algorithmic}
\usepackage{xspace}
\usepackage{xcolor}
\usepackage{paralist}
\usepackage{bm}
\usepackage{bbm}
\usepackage{amssymb}
\usepackage{amsfonts} 
\usepackage{amsmath} 
\usepackage{caption}
\usepackage{subcaption}
\usepackage{booktabs}
\usepackage{pifont}
\usepackage{enumitem}
\usepackage[export]{adjustbox}
\usepackage{makecell}
\usepackage{footmisc}

\definecolor{dodgerblue}{rgb}{0.12,0.565,1}
\newcommand{\method}{{\sc HPOD}\xspace}
\newcommand{\methodz}{{\sc HPOD\_0}\xspace}
\newcommand{\ftrain}{{PPE}\xspace}
\newcommand{\ffree}{{MSF}\xspace}

\newcommand{\f}{{$f(\cdot)$}\xspace}
\newcommand{\s}{{$s(\cdot)$}\xspace}
\newcommand{\aq}{{$a(\cdot)$}\xspace}

\newcommand{\vlambda}{{\bm{\lambda}}}
\newcommand{\vLambda}{{\bm{\Lambda}}}

\newcommand{\argmax}{\operatornamewithlimits{argmax}}

\newcommand{\cbit}{\begin{compactitem}}
	\newcommand{\ceit}{\end{compactitem}}
\newcommand{\cben}{\begin{compactenum}}
	\newcommand{\ceen}{\end{compactenum}}
	
\newtheorem{problem}{Problem}

\newcommand{\R}{\mathbb{R}}

\newcommand{\mM}{\bm{\mathcal{M}}}

\newcommand{\bXs}{\mathbf{X}_{\text{test}}}
\newcommand{\bX}{\mathbf{X}}
\newcommand{\by}{\mathbf{y}}

\newcommand{\bP}{\mathbf{P}}

\newcommand{\bD}{\mathbf{D}}
\newcommand{\bM}{\mathbf{M}}

\newcommand{\mDt}{\bm{\mathcal{D}}_{\text{train}}}
\newcommand{\mDs}{\bm{\mathcal{D}}_{\text{test}}}
\newcommand{\mD}{\mathcal{D}}

\newcommand{\bmi}{\mathbf{m}_{i}}
\newcommand{\bmt}{\mathbf{m}_{\text{test}}}

\newcommand{\bIij}{\mathbf{I}_{i,j}}

\newcommand{\bItk}{\mathbf{I}_{\text{test},k}}

\newcommand{\mO}{\mathcal{O}}

\newcommand{\algrule}[1][.5pt]{\par\vskip.5\baselineskip\hrule height #1\par\vskip.5\baselineskip}
\makeatother

\begin{document}

\newcommand\relatedversion{}


\title{Hyperparameter Optimization for Unsupervised Outlier Detection}
\author{Yue Zhao\thanks{Carnegie Mellon University, H. John Heinz III College. Email: zhaoy@cmu.edu, lakoglu@andrew.cmu.edu}
\and Leman Akoglu\footnotemark[1]}

\date{}

\maketitle







\begin{abstract} \small\baselineskip=9pt 

Given an unsupervised outlier detection (OD) algorithm, how can we optimize its hyperparameter(s) (HP) on a \textit{new dataset, without any labels}? In this work, we address this challenging hyperparameter optimization for unsupervised OD problem, and propose \textit{the first systematic approach} called \method that is based on meta-learning. \method capitalizes on the prior performance of a large collection of HPs on \textit{existing} OD benchmark datasets, and transfers this information to enable HP evaluation on a new dataset without labels. 
Also, \method adapts a 
prominent sampling paradigm to identify promising HPs efficiently. 
Extensive experiments show that \method works for both deep (e.g., Robust AutoEncoder) and shallow (e.g., Local Outlier Factor (LOF) and Isolation Forest (iForest)) OD algorithms on discrete and continuous HP spaces, and outperforms a wide range of baselines with on average 58\% and 66\% performance improvement over the default HPs of LOF and iForest.
\end{abstract}

\section{Introduction}
\label{sec:intro}
Although a long list of unsupervised outlier detection (OD) algorithms have been proposed in the last few decades \cite{Aggarwal2013a,Campos2016evaluation,han2022adbench,pang2021deep}, how to optimize their hyperparameter(s) (HP) remains underexplored. Without hyperparameter optimization (HPO) methods, practitioners often use the default HP of an OD algorithm, which is hardly optimal given many OD algorithms are sensitive to HPs. For example, a recent study by Zhao \textit{et al.} reports that by varying the number of nearest neighbors in local outlier factor (LOF) \cite{Breunig2000a} while fixing other conditions, up to 10$\times$ performance difference is observed in some datasets \cite{zhao2021automatic}. The literature also shows that HP sensitivity is exacerbated for deep OD models with more `knobs' (e.g., HPs and architectures) 
\cite{ding2022hyperparameter}, which we also observe in this study---deep robust autoencoder (RAE) \cite{zhou2017anomaly} exhibits up to 37$\times$ performance variation under different HPs in several datasets.


\begin{figure*} [!t]
    \centering
	\subfloat[(left) large perf. variation over 384 HPs for \textbf{RAE} on \texttt{Thyroid}---random\&min. loss HPs are sub-par; (right) \method (\textcolor{dodgerblue}{\textbf{\textit{H}}}) outperforms all.] {%
			\includegraphics[clip,width=0.232\columnwidth]{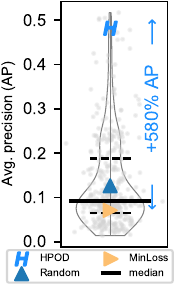}%
		\includegraphics[clip,width=0.44\columnwidth]{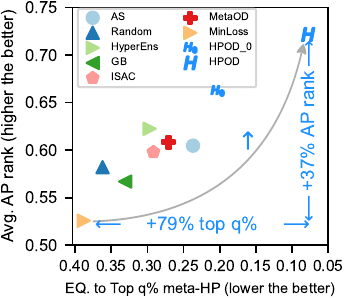}%
		\label{fig:rae_demo}
	}
	\hspace{0.05in}
	\subfloat[(left) HP sensitivity of \textbf{LOF} on \texttt{Vowels}---the default HP is far from optimal; (right) \method (\textcolor{dodgerblue}{\textbf{\textit{H}}}) outperforms all baselines with +58\% AP rank over random HP.]{%
		\includegraphics[clip,width=0.395\columnwidth]{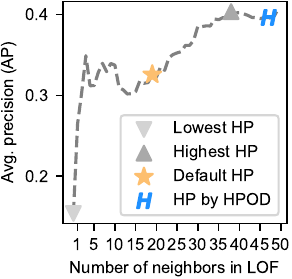}%
		\includegraphics[clip,width=0.44\columnwidth]{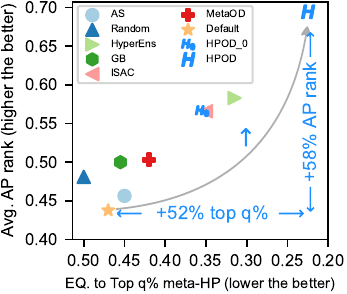}%
	\label{fig:lof_demo}
	}
	\hspace{0.05in}
	\subfloat[Perf. comparison on ensemble \textbf{iForest}; \method (\textcolor{dodgerblue}{\textbf{\textit{H}}}) outperforms all baselines markedly.] {%
		\includegraphics[clip,width=0.432\columnwidth]{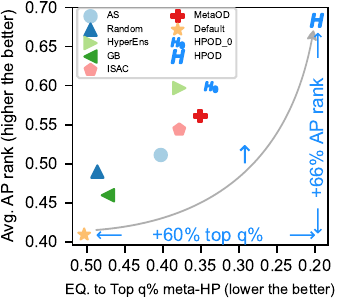}%
		\label{fig:iforest_demo}
	}
	\vspace{-0.1in}
	\caption{
	(\textit{a}) (left) Demo of HP sensitivity in deep RAE on \texttt{Thyroid}; (right) \method 
	outperforms
	all baselines on a 39-dataset database (see \S \ref{exp:rae}), with a higher avg. performance rank (y-axis) as well as comparable to better top q\% HP settings in the meta-HP set (x-axis) (\textit{b}) (left) HP sensitivity in LOF on \texttt{Vowels}; (right) \method 
	outperforms all baselines with great performance edge (e.g., $+58\%$ normalized AP rank) over the default HP (\S \ref{exp:lofif})
and (\textit{c}) for iForest, \method is also the best with huge improvement (e.g., $+66\%$ normalized AP rank) over the default HP (\S \ref{exp:lofif}). See detailed experiment results in \S \ref{sec:exp}.
	}
	\vspace{-0.2in}
	\label{fig:overall}
\end{figure*}

In supervised learning, one can use 
ground truth labels to evaluate the performance of an HP, leading to a simple grid and random search \cite{bergstra2012random} and more
efficient
Sequential Model-based Bayesian Optimization (SMBO) \cite{DBLP:journals/jgo/JonesSW98}. Unlike the simple methods, SMBO builds a cheap regression model (called the surrogate function) of the expensive objective function (which often requires ground truth labels), and uses it to iteratively select the next promising HP for the objective function to evaluate. Notably, learning-based SMBO is often more efficient than simple, non-learnable methods
\cite{falkner2018bohb,thornton2013auto}. 

However, unsupervised OD algorithms face evaluation challenges---they do not have access to (external) ground truth labels, and 
most of them 
(e.g., LOF and Isolation Forest (iForest) \cite{liu2008isolation}) do not have an (internal) objective function to guide the learning either.
Even for 
the OD algorithms 
with an internal objective (e.g., reconstruction loss in RAE), 
its value does not necessarily correlate with the actual detection performance \cite{ding2022hyperparameter}. 
Thus, HPO for unsupervised OD remains underexplored, where the key is reliable model evaluation.



Moving beyond proposing another detection algorithm, we study this important \underline{H}yper\underline{P}arameter \underline{O}ptimization for unsupervised \underline{OD} 
problem,
and introduce (to our best knowledge) \textit{the first 
solution} called \method.
In a nutshell, \method leverages meta-learning to enable (originally supervised) SMBO for efficient unsupervised OD hyperparameter optimization.
To overcome 
the infeasibility of evaluation in 
unsupervised OD, \method uses meta-learning that carries over past experience on prior datasets/tasks to more efficient learning on a new task. To that end, we build a meta-database with historical performances of a large collection of HPs on an extensive corpus of existing OD benchmark datasets, 
and train a \textit{proxy performance evaluator} (\ftrain) 
to evaluate HPs on a new dataset \textit{without labels}. 
With \ftrain,
\method can iteratively and efficiently identify promising HPs to evaluate and output the best.
Additionally, we use meta-learning to further facilitate 
\method 
by \textit{initializing} and \textit{transferring knowledge} from similar historical tasks to the surrogate function of the new task.
We find it important to remark that \method is strictly an HPO technique other than a new detection algorithm.

\textbf{Performance}. Fig. \ref{fig:rae_demo} (left) shows the huge performance variation (up to 37$\times$) for a set of 384 deep RAE HPs on \texttt{Thyroid} data, where \method is  significantly better than 
expectation (i.e. random selection),
as well as selection by min. reconstruction loss (MinLoss); ours is one of the top HPs. 
In Fig. \ref{fig:rae_demo} (right), we show that \method is significantly better than a group of diverse and competitive baselines (see Table \ref{table:baseline}) on a 39 dataset database. 
We also demonstrate \method's generality on non-deep OD algorithm LOF with both discrete and continuous HP spaces in Fig. \ref{fig:lof_demo}, as well as the 
popular iForest in Fig. \ref{fig:iforest_demo}. 
For all three OD algorithms, 
\method is \textit{statistically} better than (most) baselines, including the default HPs of LOF and iForest which are widely used by practitioners. 
In fact, being an ensemble, iForest has been shown to be robust to HPs and outperform many other detectors in the literature \cite{emmott2015meta}. As such, improving over its default setting by our \method is remarkable.


We summarize the key contributions as follows:

	\cbit
	\item \textbf{First 
	HPO framework 
	for Unsupervised Outlier Detection}. We introduce \method, a novel meta-learning approach 
	that 
	capitalizes on historical OD tasks with labels to select effective HPs for a new task without any labels.
	\item \textbf{Continuous HP search}. 
	Superior to all 
	 baselines in Table \ref{table:baseline}, \method works with both discrete \textit{and} continuous HPs. 
	\item \textbf{Generality and Effectiveness}. Extensive results on 39 datasets with (\textit{a}) deep method RAE and classical methods (\textit{b}) LOF and (\textit{c}) iForest show that \method is markedly better than 
	leading baselines,  with on avg. 58\% and 66\% improvement over the default HPs of LOF and iForest.
	\ceit
We release \method and the meta-train database (\url{https://tinyurl.com/HPOD22}) for the community to use and extend with more datasets for even better performance.

\vspace{-0.05in}
\section{Related Work}
\label{sec:related}

\setlength\tabcolsep{1.5 pt}
\newcommand{\cmark}{\ding{51}}%
\newcommand{\xmark}{\ding{55}}%

\begin{table}[!h]

\caption{\method and baselines for comparison with categorization by (1st row) whether it uses meta-learning 
 and (2nd \& 3rd row) whether it supports \textit{discrete} and \textit{continuous} HPO. Only \method and 
 \methodz leverage meta-learning and support continuous HPO.
 See 
 \S \ref{subsec:exp_details}.
 } 
\vspace{-0.1in}
\centering
\scalebox{0.56}{
\begin{tabular}{l|ccc|ccccc|cc}
\toprule
\textbf{Category}      & \textbf{Default} & \textbf{Random} & \textbf{MinLoss} & \textbf{HE} & \textbf{GB} & \textbf{ISAC} & \textbf{AS} & \textbf{MetaOD} &  \textbf{\methodz} & \method \\
\midrule
\textbf{meta-learning} &    \xmark &    \xmark            &   \xmark        &  \xmark        &        \cmark  &          \cmark    &           \cmark &     \cmark & \textcolor{dodgerblue}{\cmark} & \textcolor{dodgerblue}{\cmark} \\ 
\midrule
\textbf{discrete HP} &         \xmark     &    \xmark   &       \cmark    &      \xmark    &        \cmark  &          \cmark    &           \cmark &     \cmark & \textcolor{dodgerblue}{\cmark} & \textcolor{dodgerblue}{\cmark}     \\ 
\textbf{continuous HP} &       \xmark   &    \xmark       &  \cmark         &      \xmark    &       \xmark   &     \xmark         &      \xmark      & \xmark& \textcolor{dodgerblue}{\cmark} & \textcolor{dodgerblue}{\cmark}     \\ 

\bottomrule
\end{tabular}
}
	
	\label{table:baseline} 
	\vspace{-0.25in}
\end{table}
\setlength\tabcolsep{6 pt}
\subsection{Hyperparameter Optimization (HPO) for OD.}
We can categorize the short list of HPO methods for OD into two groups.
The first group of methods require a hold-out set with ground truth labels for evaluation \cite{bahri2022automl}, including AutoOD \cite{li2021autood}, TODS \cite{lai2021revisiting}, and PyODDS \cite{li2020pyodds}, which do not apply to unsupervised OD. The second group 
uses the default HP, 
randomly picking an HP, selecting the HP by internal objective/evaluation  \cite{goix2016evaluate,marques2020internal}, or averaging the outputs of randomly sampled HPs \cite{wenzel2020hyperparameter};
we include them as baselines (see col. 2-5 of Table \ref{table:baseline}) 
with empirical results in \S \ref{subsec:exp_results}.

\vspace{-0.075in}
\subsection{Hyperparameter Opt. and Meta-Learning.}
\label{subsec:hpo_meta_learning}
\vspace{-0.025in}

HPO gains great attention due to its advantages in searching and optimizing through complex HP spaces, where learning tasks are costly \cite{karmaker2021automl}. Existing methods include simple grid and random search \cite{bergstra2012random} and more efficient Sequential Model-based Bayesian Optimization (SMBO) \cite{DBLP:journals/jgo/JonesSW98}. Notably, SMBO builds a cheap regression model (called the surrogate function) of the expensive objective function, and uses it to iteratively select the next promising HPs to be evaluated by the objective function (see Appx. \ref{appx:SMBO}). 
We cannot directly use these supervised methods, 
rather, \method leverages meta-learning to enable efficient 
SMBO 
for \textit{HPO for OD}.


Meta-learning aims to facilitate new task learning by leveraging and transferring knowledge from prior/historical tasks \cite{vanschoren2018meta}, which has been used in warm-starting \cite{feurer2014using} and transferring surrogate \cite{wistuba2016two,wistuba2018scalable,yogatama2014efficient} in SMBO. Recently, meta-learning has also been applied to unsupervised outlier model selection (UOMS), where Zhao \textit{et al.} proposed MetaOD \cite{zhao2021automatic} with comparison to baselines including global best (GB), ISAC \cite{conf/ecai/KadiogluMST10}, and ARGOSMART (AS) \cite{nikolic2013simple}. We adapt these UOMS methods as baselines for HPO for OD (see col.
6-9 of Table \ref{table:baseline}). Although these methods leverage meta-learning, they cannot tackle continuous HPO. \method outperforms them in all experiments (see \S \ref{subsec:exp_results}).   
\vspace{-0.1in}

\section{\method: Hyperparameter Optimization for Unsupervised Outlier Detection}
\label{sec:method}
\subsection{Problem Statement}
We consider the hyperparameter optimization (HPO) problem for unsupervised outlier detection (OD), referred to as 
\textit{HPO for OD} hereafter. Given a new dataset $\bD_{\text{test}} = (\bXs, \emptyset)$ \textit{without any labels} and an OD algorithm $M$ with the HP space $\vLambda$, the goal is to identify a HP setting $\vlambda \in \vLambda$ so that model $M_\vlambda$ (i.e., detector $M$ with HP $\vlambda$) achieves the highest performance\footnote{\label{footnote:metric}In this paper, we use the area under the precision-recall curve (AUCPR, a.k.a. Average Precision or AP) as the
performance metric, which can be substituted with any other metric of interest.}. 
HPs can be discrete and continuous, leading to an infinite number of candidate HP configurations. For instance, given $h$ hyperparameters $\lambda_1 \dots \lambda_h$, with domains $\Lambda_1, \dots, \Lambda_h$, the hyperparameter space $\vLambda$ of $M$ is a subset of the cross product
of these domains: $\vLambda \subset \Lambda_1 \times \dots \times \Lambda_h$. Eq. \eqref{eq:hp_opt} presents the goal formally.

\vspace{-0.15in}
\begin{equation} \label{eq:hp_opt}
\argmax_{\vlambda \in \vLambda} \;\;
\text{perf}
(M_\vlambda, \bXs)
\end{equation}

\vspace{-0.15in}
\begin{problem}
[HPO for OD]
\textit{\em Given} a new input dataset (i.e., detection task\footnote{Throughout text, we use \textit{task} and \textit{dataset} interchangeably.}) $\mDs = (\bXs, \emptyset)$  without any labels, 
\textit{\em pick} a hyperparameter setting $\vlambda \in \vLambda$ for a given detection algorithm $M$ to employ on $\bXs$ to maximize its performance.
\end{problem}
\vspace{-0.05in}

\noindent It is infeasible to evaluate an infinite number of configurations with continuous HP domain(s), and thus
a key challenge is efficiently searching the space.
As \textit{HPO for OD} does not have access to ground truth labels $\mathbf{y}_{\text{test}}$,
HP performance (perf.)
cannot be evaluated directly.
\vspace{-0.1in}

\subsection{Overview of \method.}
\label{subsec:overview}
\vspace{-0.05in}


In \method, we use \textit{meta-learning} to enable (originally supervised) Sequential Model-based Bayesian Optimization for efficient \textit{HPO for OD}, where the key idea is to transfer useful information from historical tasks to a new test task. As such, \method takes as input a collection of historical tasks 
$\mDt = \{\mD_1,\ldots,\mD_n\}$, namely, a meta-train database with ground-truth labels where $\{\mD_i = (\bX_i,\by_i)\}_{i=1}^n$. Given an OD algorithm $M$ for HPO, we define a finite meta-HP set 
by discretizing continuous HP domains (if any) to get their cross-product, i.e.,  
$\vlambda_{\mathbf{meta}} = \{\vlambda_1, \ldots, \vlambda_m \} \in \vLambda$. We use $M_j$ to denote detector $M$ with the $j$-th HP setting $\vlambda_j \in \vlambda_{\mathbf{meta}}$.
\method uses $\mDt$ to compute:
\cbit
\item the historical output scores of each detector $M_j$
on each meta-train dataset $\mD_i \in \mDt$, where $\mO_{i,j} := M_j(\mD_i)$ refers to the output outlier scores using the $j$-th HP setting for the points in the $i$-th meta-train dataset $\mD_i$; 
and 
\item the historical performance matrix $\bP\in \R^{n\times m}$  
of each detector $M_j$,
where $\bP_{i,j}:=\text{perf}(\mO_{i,j})$ is $M_j$'s performance\footref{footnote:metric}
on meta-train dataset $\mD_i$. 
\ceit

\method consists of two phases. During \textbf{the (offline) meta-learning}, it leverages meta-train database with labels to build a \textit{proxy performance evaluator} (\ftrain), which can predict HP performance 
on a new task without labels. Also, it trains a \textit{meta-surrogate function} (\ffree) for each meta-train dataset to facilitate later HPO on a new dataset. In the \textbf{(online) HP Optimization} for a new task, \method uses \ftrain to predict its HPs' performance without using any labels, under the SMBO framework to identify promising HP settings in iteration effectively. Also, we improve the surrogate function 
in \method 
by transferring knowledge 
from \textit{similar} meta-train datasets.
An outline of \method is given in Algo. \ref{algo:hpod}, where we elaborate on the details of offline meta-training and online \textit{HPO for OD} in \S \ref{subsec:meta_train} and \S \ref{subsec:online_hpo}.

\begin{algorithm}[!t]
	\caption{\method: Offline and Online Phases}
	\label{algo:hpod}
	\small{
	\begin{algorithmic}[1]
		\REQUIRE (Offline) meta-train database $\mDt = \{(\bX_i, \by_i)\}_{i=1}^n$, OD algorithm $M$, 
		meta-HP set $\vlambda_{\mathbf{meta}} = \{\vlambda_1, \ldots, \vlambda_m \} \in \vLambda$, 
		performance evaluation $\text{perf}(\cdot)$;
		(Online) new OD dataset $\mDs=(\bXs, \emptyset)$ (no labels), 
		number of iterations $E$
		\ENSURE (Offline) \method meta-learners; (Online) the selected hyperparameter setting $\vlambda^{*}$ for $\mDs$
		\algrule
		\textcolor{dodgerblue}{$\blacktriangleright$ (Offline) \textbf{Meta-train}: 
		\textit{Learn functions for HP performance prediction}
		}
		{{\textcolor{dodgerblue}{\ttfamily{\bf (\S \ref{subsec:meta_train})}}}} 
		\STATE Train detector $M$ with
		HP 
		$\vlambda_j \in \vlambda_{\mathbf{meta}}$ on 
		each $\bX_i$ of $\mD_i \in \mDt$ to get outlier scores $\mO_{i,j}$,  $\forall\; i=1\ldots n, \; j=1\ldots m$
		\STATE Evaluate each $\mO_{i,j}$ against 
  true
  labels $\by_i$ to get performance matrix $\bP \in \R^{n\times m}$, where 
		$\bP_{i,j} := \text{perf}(\mO_{i,j} | \by_i)$

		\STATE Extract meta-features (MF) per task, $\bmi := \psi(\bX_i)$ 
		\STATE Compute internal perf. measures (IPM), $\bIij := \phi(\mO_{i,j})$ 

		\STATE Train \textit{proxy performance evaluator} (\ftrain) $f(\cdot)$ to predict the performance $\bP_{i,j}$ from the respective \{HP Settings
		$\vlambda_j$, meta-features $\bmi$, IPMs $\bIij$\} 
		\COMMENT{\S \ref{subsec:f_train}}
		\STATE Train each \textit{meta-surrogate function} (\ffree) $t(\cdot)$ per meta-train dataset $ \bm{\mathcal{T}} = \{t_1, \ldots, t_n\}$ to predict the perf. $\bP_{i,j}$ from only the respective \{HP setting $\vlambda_j$\} \COMMENT{\S \ref{subsec:MSF}}
		\STATE \textbf{Save} MF extractor $\psi$,  IPM extractor $\phi$, \ftrain $f$, and \ffree $\bm{\mathcal{T}}$
		\algrule
		\textcolor{dodgerblue}{$\blacktriangleright$ (Online) \textbf{HPO on a new task}: \textit{Iteratively identify promising HP settings and output the best one}} {{\textcolor{dodgerblue}{\ttfamily{\bf (\S \ref{subsec:online_hpo})}}}}
		\STATE Extract meta-features of the test task $\bmt := \psi(\bXs)$ 

		\STATE Initialize surrogate function $s^{(1)}$ and the evaluation set $\vlambda_{\text{eval}}$ by the meta-train database and \ftrain \COMMENT{\S \ref{subsec:init}}
		\FOR[\S \ref{subsec:use_smbo}]{$e=1$ to $E$} 
		\STATE Transfer meta-surrogate functions $\bm{\mathcal{T}}$ to surrogate $s^{(e)}$ by performance similarity to meta-train
		\COMMENT{\S \ref{subsec:surrogate_transfer}}
		\STATE Get the promising HP 
		to evaluate by 
		EI on surrogates' prediction, 
		$\vlambda^{(e)} := \argmax_{\vlambda_{k} \in \vlambda_{\mathbf{sample}}} EI(\vlambda_{k}|s^{(e)})$
		\STATE Build $M$ with $\vlambda^{(e)}$, and get the corresponding outlier scores $\mO_{\text{test}}^{(e)}$ and IPMs $\bm{\mathcal{I}}_{\text{test}}^{(e)}$
        
		\STATE Predict performance of 
		$\vlambda^{(e)}$ 
		 by \f,
		$\widehat{\mathbf{P}}_{\text{test}}^{(e)} :=f(\vlambda^{(e)}, \bmt, \bm{\mathcal{I}}_{\text{test}}^{(e)})$
		\STATE Add $\vlambda^{(e)}$ to the evaluation set $\vlambda_{\text{eval}} := \vlambda_{\text{eval}} \cup \vlambda^{(e)}$ 
		\STATE Update to $s^{(e+1)}$ with new pairs of info. $\langle\vlambda^{(e)},\widehat{\mathbf{P}}_{\text{test}}^{(e)}\rangle$
		\ENDFOR
		\STATE \textbf{Output} $\vlambda^{*} \in \vlambda_{\text{eval}}$ w/ the highest predicted perf. 
	\end{algorithmic}
	}
\vspace{-0.05in}
\end{algorithm}
 \setlength{\textfloatsep}{0.1in}

\vspace{-0.05in}
\subsection{(Offline) Meta-training} 
\label{subsec:meta_train}
\vspace{-0.05in}

In principle, meta-learning carries over the prior experience of historical (meta-train) tasks to do better on a new task, given the latter at least resembles some of the historical tasks. 
Due to the lack of ground truth labels and/or a reliable internal objective function,  the key challenge in HPO for OD is to evaluate the performance of HP settings. Thus, the core of \method's meta-learning is \textit{learning} the mapping from \textit{HP settings} onto \textit{ground-truth performance} by the \textit{supervision} from the meta-train database. The first part (lines 1-7) of Algo. \ref{algo:hpod} describes the core steps, and we further discuss on how to learn this mapping (\S \ref{subsec:f_train}) and transfer additional information for a new task (\S \ref{subsec:MSF}) in the following. Notably, \textit{offline} meta-training is \textit{one-time} and amortized over many test tasks.

\vspace{-0.1in}

\subsubsection{Proxy Performance Evaluator (\ftrain).} 
\label{subsec:f_train}
In \method, we learn a regressor \f across all meta-train datasets, named \textit{Proxy Performance Evaluator}, that maps their  
\{\texttt{HP settings}, \texttt{data characteristics}, \texttt{additional signals}\} onto ground truth performance.
If \f only uses HP settings as the input feature,
it fails to capture the performance variation of an HP given different datasets.
Thus, we need additional input features to enable \f for quantifying dataset similarity, so that \f can make similar HP performance predictions on similar datasets, and vice versa.

How can we capture dataset similarity in OD? 
Recent work by 
\cite{zhao2021automatic} introduced specialized OD meta-features (MF)  to describe general characteristics of OD datasets; e.g., number of samples, basic feature statistics, etc. With the meta-feature extractor, both meta-train datasets and (later) the test dataset can be expressed as fixed-length vectors, and thus any similarity measure applies, e.g., Euclidean distance. To build \f, we extract meta-features from each meta-train dataset as
$\bM= \{\mathbf{m}_1, \ldots, \mathbf{m}_n\}=\psi(\{\bX_1, \ldots, \bX_n\})\in \mathbbm{R}^{n \times d}$,
where $\psi(\cdot)$ is the extraction module, and $d$ is the dimension of meta-features (see \cite{zhao2021automatic} for details).

Although meta-features describe general characteristics of OD datasets, their similarity does not necessarily correlate with the \textit{actual performance}. Thus, we enrich the input features of \f  with internal performance measures (IPMs) \cite{ma2021large}, which are more ``performance-driven''. 
 IPMs are noisy/weak unsupervised signals that are solely based on the input samples and/or a given model's output (e.g., outlier scores)
that can be used to compare two models \cite{goix2016evaluate,marques2020internal}. 
In \method, we make the best use of these weak signals by \textit{learning} in \f to regress the IPMs of a given HP setting (along with other signals) onto its true performance \textit{with supervision}.
To build \f, we extract IPMs of 
each detector $M_j$ with HP setting $\vlambda_j \in \vlambda_{\mathbf{meta}}$ on each meta-train dataset $\mD_i \in \mDt$, where $\bIij := \phi(\mO_{i,j})$ refers to the IPMs using the $j$-th HP setting
for the $i$-th meta-train dataset, and $\phi(\cdot)$ is the IPM extractor. We defer the details on IPMs to Appx. \ref{appx:ipm}.

Putting these together, 
we build \textit{Proxy Performance Evaluator} \f to
map 
\{\texttt{HP setting}, \texttt{meta-features}, \texttt{IPMs}\} of HP $\vlambda_j \in \vlambda_{\mathbf{meta}}$ on the $i$-th meta-train dataset onto its 
ground truth performance, i.e.,
$f(\vlambda_j, \mathbf{m}_i, \bIij) \mapsto \bP_{i,j}$. 
See 
details of \f in Appx. \ref{appx:ftrain}.

We find it critical to remark that provided $\psi(\cdot)$, $\phi(\cdot)$, and the trained \f  at test time, predicting the detection performance of HP settings becomes possible for the new task without using any ground-truth labels. 
\vspace{-0.25in}

\subsubsection{Meta-Surrogate Functions (\ffree).} 
\label{subsec:MSF}
Different from \f  that trains on \textit{all} meta-train datasets and leverages \textit{rich input features} (i.e., HPs, MFs, and IPMs) to predict HP performance, we also train $n$ \textit{independent} regressors with \textit{only HPs as input}, $\bm{\mathcal{T}} = \{t_1, \ldots, t_n\}$. That is, for each meta-train dataset $\mD_i \in \mDt$, we train a regressor $t_i(\cdot)$ that simply maps the $j$-th HP setting $\vlambda_j \in \vlambda_{\mathbf{meta}}$ to its detection performance on the $i$-th meta-train dataset, i.e., $t_i(\vlambda_j) \mapsto \bP_{i,j}$.

Since these independent
regressors only use HP settings as input, 
they can be readily transferred to the online model selection phase to improve HP performance evaluation on $\bXs$. We defer their specific usage to \S \ref{subsec:use_smbo} and \S \ref{subsec:surrogate_transfer}.
\vspace{-0.05in}

\subsection{(Online) HPO on a New OD Task.}
\label{subsec:online_hpo}
\vspace{-0.1in}

After the meta-training phase, \method is ready to optimize HPs for a new dataset. In short, it outputs the HP with the highest predicted performance by \f, the trained performance evaluator (\S \ref{subsec:init}). 
To explore 
better HPs efficiently,
\method leverages Sequential Model-based Optimization to iteratively select promising HPs for evaluation (\S \ref{subsec:use_smbo}). The second part (lines 8-18) of Algo. \ref{algo:hpod} shows the core steps.
\vspace{-0.1in}

\subsubsection{Hyperparameter Optimization via Proxy Performance Evaluator.} 
\label{subsec:init}
Given a new dataset $\mDs$, we can sample 
a set of 
HPs (termed as the evaluation set $\vlambda_\text{eval} \in \vLambda$), and use the \textit{proxy performance evaluator} \f from meta-training to predict their performance, based on which we can output the one with the highest predicted value as follows.

\vspace{-0.18in}
\begin{equation}
\vspace{-0.05in}
    \argmax_{\vlambda_k \in \vlambda_{\text{eval}}} \; f(\vlambda_k, \bmt, \bItk)
    \label{eq:selected_hp}
    \vspace{-0.05in}
\end{equation}

By setting $\vlambda_{\text{eval}}$ to some \textit{randomly sampled HPs} and plugging it into Eq. (\ref{eq:selected_hp}), 
we have the ``version 0" of \method, referred as \methodz.
However, \f  needs IPMs (i.e., $\bItk$ in Eq. (\ref{eq:selected_hp})) as part of the input, requiring detector building at test time. 
Thus, we should construct $\vlambda_{\text{eval}}$ carefully to ensure it captures promising HPs, where random sampling is insufficient.
Thus we ask: how can we efficiently identify promising HPs for model building and evaluation at  test time? 
\vspace{-0.08in}

\subsubsection{Identifying Promising HPs 
by Sequential Model-based Optimization (SMBO).}
\label{subsec:use_smbo}
As we briefly described in \S \ref{sec:related}, SMBO can iteratively optimize an expensive objective \cite{hutter2011sequential}, and has been widely used in supervised model selection and HPO \cite{bergstra2015hyperopt}. Other than sampling HPs randomly, learning-based SMBO shows better efficiency in finding promising HPs to evaluate in iterations. 
In short, 
SMBO constructs a cheap regression model (called surrogate function \s) and uses it for identifying the promising HPs to be evaluated by the (expensive) true objective function. It then iterates between fitting the surrogate function with newly evaluated HP information and gathering new information based on the surrogate function. We provide the pseudo-code of the \textit{supervised} HPO by SMBO in Appx. Algo. \ref{algo:smbo}, and note that it does not 
directly 
apply to \textit{HPO for OD} as perf. cannot be evaluated without ground truth labels (line 4).

We enable (originally \textit{supervised}) SMBO for \textit{unsupervised} outlier detection HPO by plugging the \ftrain \f from the meta-train in place of HP performance evaluation. The key steps of online HPO 
are presented below.

\noindent \textbf{\textit{Surrogate Function and Initialization}}. As an approximation of the expensive objection function, surrogate function \s only takes HP settings as input, aiming for fast performance evaluation on a large collection of sampled HPs.
For the new task $\bXs$ without 
access to true performance evaluation,
\method lets \s learn a mapping from an HP $\vlambda_k$ to its predicted performance, 
i.e.,  $s(\vlambda_k) \mapsto f(\vlambda_k, \mathbf{m}_\text{test}, \bItk)$. To enable \f on $\bXs$, \method needs one-time computation for the 
corresponding 
meta-features as $\mathbf{m}_{\text{test}}:=\psi(\bXs) \in \mathbbm{R}^{d}$. 
We want to remark that \s differs from \f in two aspects. 
First, \s can make fast performance predictions on HPs as it only needs HPs as input, while \f is more costly since IPMs 
require model building.
Second, \s is a regression model that can measure both \textit{the predicted performance} of HP settings and 
\textit{uncertainty (potential) around the prediction} simultaneously. 
A popular choice for \s is the Gaussian Process (GP)  \cite{williams1995gaussian}\footnote{\label{footnote:gp}We use GP in \method; one may use any regressor with prediction uncertainty estimation, e.g., random forests \cite{breiman2001random}.}. 

To initialize \s, we train it on a small number of HPs.
More specifically, we train \s with pairs of HPs and their corresponding predicted performance by \f on $\bXs$, and also initialize the evaluation set $\vlambda_{\text{eval}}$ to these HPs. 
Although we can randomly sample the initial HPs,
we propose to set them to top-performing HPs from similar meta-train tasks.
Consequently, our initial \s 
is more accurate in predicting 
likely well-performing HPs on $\bXs$.
We defer the details of this meta-learning-based
surrogate initialization to \S \ref{subsubsec:init}.

\noindent \textbf{\textit{Iteration: Identifying Promising HPs}.} Although we can already output an HP from $\vlambda_{\mathbf{eval}}$ with the highest predicted performance after initialization,
we aim to use \s to identify ``better and better'' HPs.

In each iteration, we use \s to predict the performance (denoted as $u_k:=s(\vlambda_k)$) and the uncertainty around the prediction (denoted as $\sigma_k$) of 
sampled $\vlambda_k \in \vlambda_{\mathbf{sample}}$\footnote{\label{footnote:sample}
$\vlambda_{\mathbf{sample}}$ is a finite HP candidate set that is randomly sampled from the full (continuous) HP space $\vLambda$ 
(see 
details in Appx. \ref{appx:sampling}).
Since \s can make fast predictions, $\vlambda_{\mathbf{sample}}$'s size can be large, e.g., 10,000 as in \cite{hutter2011sequential}.
},
and then select the most \textit{promising} one to be ``evaluated'' by \f. 
Intuitively, we would like to evaluate the HPs with both high predicted performance $u_k$ (i.e., exploitation) and 
high potential/prediction uncertainty $\sigma_k$ (i.e., exploration), which is widely known as ``exploitation-exploration trade-off'' \cite{shahriari2015taking}. Too much exploitation (i.e., always evaluating the similar HPs) will fail to identify promising HPs, while too much exploration (i.e., only considering high uncertainty HPs) may lead to low-performance HPs. Also, note that the quality of identified HPs depends on the prediction accuracy of \s, where we propose to transfer knowledge from 
MSF
$\bm{\mathcal{T}}$ by performance similarity (see technical details in \S \ref{subsec:surrogate_transfer}.)

How can we effectively balance the trade-off between exploitation and exploration in \method? 
Borrowing the idea of SMBO, 
we use the acquisition function \aq to factor in the trade-off and pick a promising HP setting based on the outputs of the surrogate function. 
The acquisition function 
quantifies the ``expected utility'' of HPs by balancing their predicted performance and the uncertainty.
Thus, we output the most promising HP
to evaluate by maximizing \aq:

\vspace{-0.1in}

\begin{equation}
    \vlambda 
    := \argmax_{
    \vlambda_k \in 
    \vlambda_{\mathbf{sample}}
    } \;\;
  a(s(\vlambda_k))
        \vspace{-0.05in}
    \label{eq:acqusition}
\end{equation}

One of the most prominent choices of \aq is Expected Improvement (EI) \cite{DBLP:journals/jgo/JonesSW98}, which is used in \method and can be replaced by other choices. EI has a closed-form expression
under the Gaussian assumption, and the EI value of HP setting $\vlambda_k$ is shown below.

\vspace{-0.15in}
\begin{equation*}
\label{eq:EI}
    EI(s(\vlambda_k))
    :=\sigma_{k} \cdot [u_{k} \cdot
    \Phi(u_{k})+\varphi(u_{k})], \quad \text{where}
    \vspace{-0.25in}
\end{equation*}
\begin{equation}
 u_{k}= 
\begin{cases}
    \frac{u_{k}-\widehat{\mathbf{P}}_\text{test}^{*}}{\sigma_{k}}  & \text{if } \sigma_{k} > 0 
    \textrm{ and } 
    \bigg\{ 0 
    \quad 
    \text{if } \sigma_{k} = 0
\end{cases}
\vspace{-0.03in}
\end{equation}

In the above, $\Phi(\cdot)$ and $\varphi(\cdot)$ respectively denote the cumulative distribution and the probability density functions of a standard Normal distribution, $u_{k}$ and $\sigma_{k}$ are the predicted performance and the uncertainty around the prediction of $\vlambda_k$ by the surrogate function \s, and $\widehat{\mathbf{P}}_\text{test}^{*}$ is the highest predicted performance by \f on $\vlambda_{\text{eval}}$  so far. We compare EI with other selection criteria in \S \ref{exp:aquisition}.

At the $e$-th iteration, we plug the surrogate $s^{(e)}(\cdot)$ into Eq. (\ref{eq:acqusition}), which returns $\vlambda^{(e)}$ to evaluate. 
Next, we train the OD model $M$ with $\vlambda^{(e)}$ to get its outlier scores $\mO_{\text{test}}^{(e)}$ and IPMs $\bm{\mathcal{I}}_{\text{test}}^{(e)}$, and then predict its perf. by \f 
\vspace{-0.1in}
\begin{equation}
\vspace{-0.05in}
    \widehat{\mathbf{P}}_{\text{test}}^{(e)} :=f(\vlambda^{(e)}, \bmt, \bm{\mathcal{I}}_{\text{test}}^{(e)})
\end{equation}
Finally, we add  $\vlambda^{(e)}$ to the evaluation set
$\vlambda_{\text{eval}} := \vlambda_{\text{eval}} \cup \vlambda^{(e)}$, and update the surrogate function to $s^{(e+1)}$ with newly evaluated HP information
$\langle\vlambda^{(e)},\widehat{\mathbf{P}}_{\text{test}}^{(e)}\rangle$.

To sum up, \method alternates between (\textit{i}) identifying the next promising HP 
by the surrogate function \s and (\textit{ii}) updating \s based on newly evaluated HP.

\noindent \textbf{\textit{Continuous HP search}}.
Recall that an outstanding property of \method compared to 
the baselines
is its capability for continuous HP search (Table \ref{table:baseline}). 
$\vlambda_{\mathbf{sample}}$\footref{footnote:sample} can be \textit{any} subset of the full HP space $\vLambda$ and not restricted to the discrete $\vlambda_{\mathbf{meta}}$. 

\noindent \textbf{\textit{Time Budget}}.
\method is an \textit{anytime} algorithm: at any time the user asks for a result, it can always output the HP with the highest predicted performance in the evaluation set $\vlambda_{\text{eval}}$ at the current iteration (Eq. (\ref{eq:selected_hp})). \method uses $E$ to denote the max number of iterations.
\vspace{-0.25in}

\subsection{Details of (Online) HPO}
\label{subsec:online_hpo_details}
\subsubsection{Meta-learning-based Surrogate Initialization.} \vspace{-0.15in}
\label{subsubsec:init}
Other than initializing on a small set of \textit{randomly sampled} HPs, we design a meta-learning initialization strategy for the surrogate function (see \S \ref{subsec:use_smbo}), based on the similarity between the test and meta-train datasets. Intuitively, we hope the surrogate function \s can make accurate predictions on the top-performing HPs of the test dataset, while the accuracy of the under-performing HP regions is less important. To this end, we use the meta-features (see \S \ref{subsec:f_train}) to calculate the similarity between the test dataset to each meta-train task $\mD_i \in \mDt$, and initialize \s with the top performing HPs from \textit{the most similar meta-train dataset}; these HPs may yield good performance on the test dataset.
See
the comparison to random initialization in \S \ref{subsec:exp_init}.
\vspace{-0.1in}

\subsubsection{Surrogate Transfer by Performance Similarity.} 
\label{subsec:surrogate_transfer}
As introduced in \S \ref{subsec:hpo_meta_learning}, meta-learning can be used to improve the surrogate function \s in SMBO by transferring knowledge from meta-train datasets. In \method, we train $n$ independent \textit{Meta-Surrogate Functions} (see \S \ref{subsec:MSF}) for $n$ meta-train datasets, $\bm{\mathcal{T}} = \{t_1, \ldots, t_n\}$; each with the same regression function as \s (i.e., Gaussian Process). To this end, we identify the most similar meta-train dataset $\mD_i$ in iteration $e$, and use the test surrogate \s and $\mD_i$'s meta-surrogate $t_i$ together to predict the performance of HP $\vlambda_k$,
i.e., 

\vspace{-0.15in}
\begin{equation}
u_k :=  s^{(e)}(\vlambda_k)+w_i^{(e)}\cdot t_i(\vlambda_k)
\vspace{-0.1in}
\end{equation}

\noindent
where $w_i^{(e)}$ 
is the similarity between the test dataset and $\mD_i$ measured in iteration $e$. While we could use meta-features
to measure the dataset similarity, its value does not change in iterations and 
finds the same meta-train dataset to transfer 
(even for different OD algorithms).

Instead, we 
(re)calculate the \textit{performance similarity} every iteration based on the HPs in $\vlambda_\text{eval}$ between each meta-train task and the test dataset, and \textit{dynamically} transfer the most similar meta-train dataset's meta-surrogate function. 
Specifically, \method computes a rank-based similarity by weighted Kendall tau \cite{shieh1998weighted}) between each meta-train dataset's ground truth performance and the test dataset's predicted performance by \f on $\vlambda_\text{eval}$ (which gets updated in every iteration).
See the effect of surrogate transfer in \S \ref{exp:transfer}.
\vspace{-0.05in}

\section{Experiments}
\label{sec:exp}
\subsection{Experiment Setting}
\label{subsec:exp_details}
\vspace{-0.05in}

\textbf{OD Algorithms and Testbed}. We show 
the results of HPO on (\textit{a}) deep RAE in \S \ref{exp:rae}
and (\textit{b}) LOF and (\textit{c}) iForest in \S \ref{exp:lofif}. Each OD algorithm is evaluated on a 39-dataset testbed (Appx. \S \ref{appx:datasets}, Table \ref{table:datasets}).
Details of each algorithm's HP spaces and the meta-HP set is provided in Appx. \ref{appx:od_details}. 
Experiments are conducted on an AMD 5900x@3.7GhZ, 64GB RAM workstation with an NVIDIA RTX A6000.

\noindent \textbf{Baselines}. Table \ref{table:baseline} summarizes the baselines
with categorization. We include
(\textit{i}) \textbf{\textit{Simple methods}}: \textbf{(1) Default} always employs the same default/popular HP setting (only if specified in the literature) \textbf{(2) Random} choice of HPs and \textbf{(3) MinLoss} outputs the HP with the lowest internal loss (only applicable to the algorithms with an objective/loss function) and (\textit{ii}) \textbf{\textit{Complex methods}}: 
\textbf{(4) HyperEnsemble (HyperEns} or \textbf{HE)} that averages the results of randomly sampled HPs \cite{wenzel2020hyperparameter}
\textbf{(5) Global Best (GB)} selects the best performing HP on meta-train database on average \textbf{(6) ISAC}  ~\cite{conf/ecai/KadiogluMST10} \textbf{(7) ARGOSMART (AS)} ~\cite{nikolic2013simple} and \textbf{(8) MetaOD} ~\cite{zhao2021automatic}.
Additionally, we include \textbf{(9) \methodz}, a variant of \method that  directly uses \f to choose from randomly sampled HPs (see \S \ref{subsec:init}). 
Note that the unsupervised OD model selection baselines (5)-(8) are not for \textit{HPO for OD}, i.e. they are infeasible with continuous HP spaces. We adapt them for HPO by selecting 
from the discrete meta-HP set in \S \ref{subsec:overview}.
See baseline details
in Appx. \ref{appx:baselines}.

\noindent \textbf{Evaluation}.  We split the meta-train/test by leave-one-out cross-validation (LOOCV). Each time we use one dataset as the input dataset for HPO, and the remaining datasets as meta-train. 
We run five independent trials and report the average for the baselines with randomness. 
We use Average Precision (AP) as the performance measure, while it can be substituted with any other measure\footref{footnote:metric}.
As the raw performance like AP is not comparable across datasets with varying magnitude, we report the normalized AP rank of an HP, ranging from 1 (the best) to 0 (the worst)---thus higher the better. Also, we provide an additional metric called ``top q\%'', denoting that an HP's performance has no statistical difference from the top q\% HP from the meta HP-set, ranging from 0 (the best) to 1 (the worst)---thus lower the better.
To compare two methods, we use the paired Wilcoxon signed rank test across all 39 datasets
(significance level $p$$<$$0.05$).
We give the full performance results in Appx. \ref{appx:full_results}.

\noindent \textbf{Hyperparameters}. The hyperparameters of \method are chosen by LOOCV on the meta-train. 
We initialize the surrogate function with 10 HPs. Also, 
we report the results within a 
30-minute budget for deep RAE and 10-minute budget for LOF and iForest.


\subsection{Experiment Results}
\label{subsec:exp_results}
\vspace{-0.05in}

\subsubsection{Results on (Deep) Robust Autoencoder.}
\label{exp:rae}
\vspace{-0.05in}

\begin{figure}[!t]
    \centering
    \vspace{-0.1in}
    \subfloat[Results on RAE]
    {
    \includegraphics[clip,width=0.8\columnwidth]{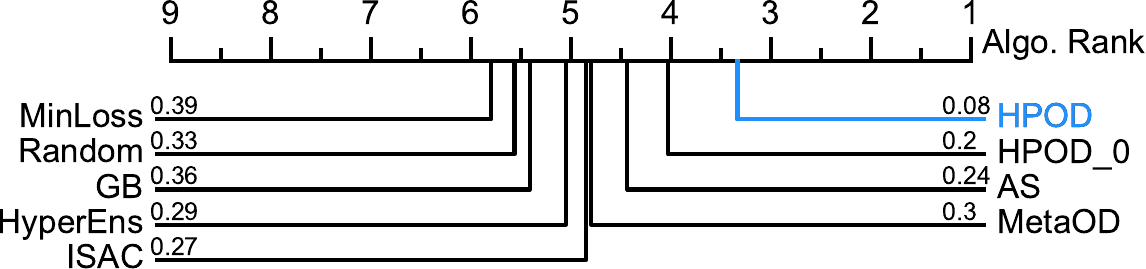}
    \label{fig:cd_plot_rae}
    }
    
    \subfloat[Results on LOF]
    {
    \includegraphics[clip,width=0.8\columnwidth]{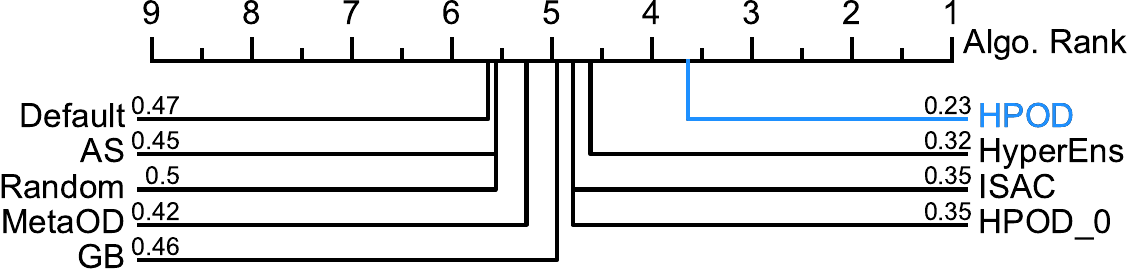}
    \label{fig:cd_plot_lof}
    }
    
    \subfloat[Results on iForest]
    {
    \includegraphics[clip,width=0.8\columnwidth]{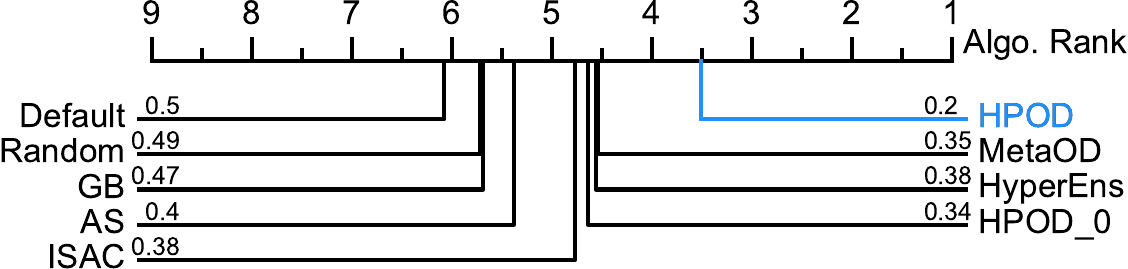}
    \label{fig:cd_plot_iforest}
    }
    \vspace{-0.1in}
    \caption{Comparison of avg. rank (lower is better) of algorithm performance across datasets on three OD algorithms. \method outperforms all w/ the lowest avg. algo. rank.
	The numbers on each line are the top q\% value (lower is better) of the employed HP (or the avg.)
	by each method. \method shows the best performance in all three HPO experiments.}
    \label{fig:cd_plots}
\end{figure}

Fig. \ref{fig:rae_demo} (right) and \ref{fig:cd_plot_rae} show that \textbf{\method outperforms all baselines w.r.t. both the best avg. normalized AP rank and the top q\% value}. Furthermore, \method is also statistically better than all baselines as shown in Table \ref{table:rae_pairs}, including strong meta-learning baseline MetaOD ($p$$=$$0.0398$). Its advantages can be credited to two reasons. First, meta-learning-based \method leverages prior knowledge on similar historical tasks to predict HP perf. on the new dataset, whereas simple baselines like Random and MinLoss cannot. Second, only \method and \methodz can select HPs from continuous spaces, while other meta-learning baselines 
are limited to finite discrete HPs as specified for the meta-train datasets
which 
are too few to capture optimal HPs (especially for deep models with huge HP spaces).

\noindent \textbf{HPO by the internal objective(s) is insufficient}. 
Fig. \ref{fig:cd_plot_rae} shows that selecting HP by minimal reconstruction loss (i.e., MinLoss) has the worst performance for RAE, even if it can work with continuous spaces. This suggests that internal loss does not necessarily correlate with external performance. On avg., \method has 37\% higher normalized AP rank, showing the benefit of transferring supervision via meta-learning.
\vspace{-0.1in}

\subsubsection{Results on LOF
and 
iForest.
}
\label{exp:lofif}
\noindent In addition to deep RAE, \textbf{\method shows generality on diverse OD algorithms}, including non-deep LOF (Fig. \ref{fig:lof_demo} and \ref{fig:cd_plot_lof}) with mixed HP spaces (see details in Appx. Table \ref{table:rae_lof})
as well as ensemble-based iForest (Fig. \ref{fig:iforest_demo} and \ref{fig:cd_plot_iforest}). \method achieves the best performance in both with the best norm. AP rank and top q\%. 

\noindent \textbf{\method is statistically better than the default HPs of LOF ($p$$=$$0.0029$) and iForest ($p$$=$$0.0013$)} (see Table \ref{table:lof_pairs} and \ref{table:iforest_pairs}). More specifically, we find that \method provides $+58\%$ and $+66\%$ performance (i.e., normalized AP rank) improvement over using the default HPs of LOF (Fig. \ref{fig:lof_demo} (right)) and iForest (Fig. \ref{fig:iforest_demo}). In fact, note that the default HPs rank the lowest for both LOF (Fig. \ref{fig:cd_plot_lof}) and iForest (Fig. \ref{fig:cd_plot_iforest}), justifying the importance of HPO methods in unsupervised OD.

\noindent \textbf{HyperEns that averages outlier scores from randomly samples HPs yield reasonable performance}, which agrees with the observations in the literature \cite{ding2022hyperparameter}. However, HyperEns has a higher inference cost as it needs to save and use all base models,
not ideal for time-critical applications. Using a single model with the selected HPs by \method offers not only better detection accuracy but also efficiency.
\vspace{-0.1in}

\begin{table}[!t]
     \caption{Pairwise statistical test results between \method and baselines by Wilcoxon signed rank test. Statistically better method shown in \textbf{bold} (both marked \textbf{bold} if no significance).}
    
    \footnotesize
    
    \label{table:all_pairs}
    \vspace{-0.1in}
    \begin{subtable}[t]{.48\textwidth}
        \centering
        \caption{On RAE, \method is statistically better than all.}\label{table:rae_pairs}
        \vspace{-0.05in}
            \scalebox{0.82}{
           \begin{tabular}{lll|lll}
            \toprule
            \textbf{Ours} & \textbf{baseline} & \textbf{p-value} & \textbf{Ours} & \textbf{baseline} & \textbf{p-value} \\
            \midrule
            \textbf{HPOD} & AS                & 0.0309           & \textbf{HPOD} & ISAC              & 0.0028           \\
            \textbf{HPOD} & Random            & 0.0014           & \textbf{HPOD} & MetaOD            & 0.0398           \\
            \textbf{HPOD} & HyperEns          & 0.0382           & \textbf{HPOD} & MinLoss           & 0.0003           \\
            \textbf{HPOD} & GB                & 0.0002           & \textbf{HPOD} & HPOD\_0            & 0.0201    \\
            \bottomrule
            \end{tabular}}
    \end{subtable}
    \vspace{0.05in}
    
    \begin{subtable}[t]{.48\textwidth}
        \centering
        \caption{On LOF, \method is statistically better than all (except HyperEnsemble (HE)), including the \textit{default} HP setting.}\label{table:lof_pairs}
                \vspace{-0.05in}
            \scalebox{0.82}{
           \begin{tabular}{lll|lll}
            \toprule
            \textbf{Ours} & \textbf{baseline} & \textbf{p-value} & \textbf{Ours} & \textbf{baseline} & \textbf{p-value} \\
            \midrule
            \textbf{HPOD} & AS                & 0.0023           & \textbf{HPOD} & ISAC              & 0.0246           \\
            \textbf{HPOD} & Random            & 0.0001           & \textbf{HPOD} & MetaOD            & 0.0088           \\
            \textbf{HPOD} & \textbf{HyperEns} & 0.0607           &   \textbf{HPOD} & Default           & 0.0029            \\
            \textbf{HPOD} & GB                & 0.0017           & \textbf{HPOD} & HPOD\_0                & 0.0016          \\
            \bottomrule
            \end{tabular}}
    \end{subtable}
    \vspace{0.05in}

    \begin{subtable}[t]{.48\textwidth}
        \centering
        \caption{On iForest, \method is statistically better than all baselines, including the \textit{default} HP setting.}\label{table:iforest_pairs}
                \vspace{-0.05in}
            \scalebox{0.82}{
           \begin{tabular}{lll|lll}
            \toprule
            \textbf{Ours} & \textbf{baseline} & \textbf{p-value} & \textbf{Ours} & \textbf{baseline} & \textbf{p-value} \\
            \midrule
\textbf{HPOD} & AS                & 0.0055           & \textbf{HPOD} & ISAC              & 0.0088           \\
\textbf{HPOD} & Random            & 0.0003           & \textbf{HPOD} & MetaOD            & 0.0289           \\
\textbf{HPOD} & HyperEns & 0.0484           &        \textbf{HPOD} & Default           & 0.0013    \\
\textbf{HPOD} & GB                & 0.0027           & \textbf{HPOD} & HPOD\_0                & 0.003                \\
            \bottomrule
            \end{tabular}}
    \end{subtable}
    \vspace{-0.1in}
   
\end{table}

\subsection{Case Study.}
\label{subsec:case_study}
\vspace{-0.05in}

\begin{table}[!htb]
\centering
	\footnotesize
		\caption{Trace of \method on \texttt{Cardiotocography} dataset. Over iterations (col. 1), \method gradually identifies better HPs (col. 2 \&3), with higher AP (col. 4). The optimal HP on from the meta-HP set is \{'Chebyshev', 79\}, which \method gets closer to the optimal HP during its adaptive search.} 
	\label{table:case_study} 
	\vspace{-0.1in}
\scalebox{0.85}{
\begin{tabular}{l|ll|l}
\toprule
\textbf{\# Iter} & \textbf{Dist. metric} & \textbf{\# Neighbors} & \textbf{AP Value} \\
\midrule
1                & Manhattan         & 23                    & 0.2866                       \\
$\ldots$ &$\ldots$&$\ldots$&$\ldots$\\
10               & Cosine           & 55                    & 0.3438                       \\
$\ldots$ &$\ldots$&$\ldots$&$\ldots$\\
20               & Chebyshev        & 72                    & 0.3569                      \\
$\ldots$ &$\ldots$&$\ldots$&$\ldots$\\
30               & Chebyshev        & 73                    & 0.357                       \\
\midrule\midrule
\textbf{Optimal} & Chebyshev        & 79                    & \textbf{0.3609}                  \\
\bottomrule          
\end{tabular}
}\vspace{-0.05in}

\end{table}

We trace how \method identifies better HPs over iterations.
We show an example of tuning LOF on \texttt{Cardiotocography} dataset in Table \ref{table:case_study}. Among 200 candidate HP settings 
(see Appx. Table \ref{table:rae_lof}), the optimal HP setting is \{`Chebyshev', 79\} with AP=0.3609. In 30 iterations, \method gradually identifies better HPs (closer to optimal), i.e., \{`Chebyshev', 73\}. Its AP improves from 0.2866 (1-\textit{st} iteration) to 0.357 (30-\textit{th} iteration). See full trace in Appx. \ref{appx:case_study}.

\subsection{Ablation Studies and Other Analysis}
\vspace{-0.1in}
\subsubsection{The Choices of Acquisition Function.}
\label{exp:aquisition}
\begin{figure}[H]
\vspace{-0.1in}
\centering
	\includegraphics[width=0.7\linewidth]{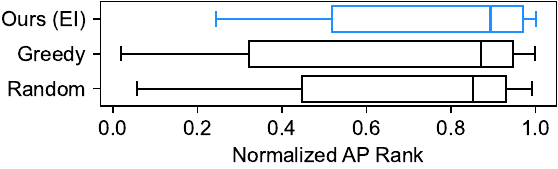}
	\vspace{-0.1in}
	\caption{Ablation of  EI (med.=0.893) vs. the greedy  (med.=0.870) and random acquisition (med.=0.851).
	}
	\label{fig:ablation_ei}
	\vspace{-0.1in}
\end{figure}

\method uses the EI acquisition to select an HP based on the surrogate function's prediction (see \S \ref{subsec:use_smbo}). We compare it with the random and greedy acquisition (latter picks the HP with the highest predicted performance, 
ignoring uncertainty) in Fig. \ref{fig:ablation_ei}, where EI-based acquisition shows the best performance.

\subsubsection{Surrogate Initialization.}
\label{subsec:exp_init}
\begin{figure}[H]
\vspace{-0.1in}
\centering
	\includegraphics[width=0.8\linewidth,left]{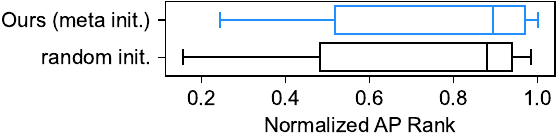}
	\vspace{-0.2in}
	\caption{Ablation of meta- (med.=0.893) vs. random-initialization (med. =0.874) of the surrogate function.
	}
	\label{fig:ablation_init}
	\vspace{-0.15in}
\end{figure}

\method uses meta-learning to initialize the surrogate (see \S \ref{subsubsec:init}). Fig. \ref{fig:ablation_init} shows its advantage over random init. with higher perf.

\subsubsection{The Effect of Surrogate Transfer.} 
\label{exp:transfer}
\begin{figure}[H]
\vspace{-0.1in}
\centering
	\includegraphics[width=0.8\linewidth,left]{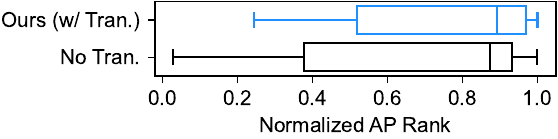}
	\vspace{-0.25in}
	\caption{Ablation of ours w/ surrogate transfer (med.=0.893) vs. without transfer (med. =0.872).
	}
	\label{fig:ablation_transfer}
	\vspace{-0.15in}
\end{figure}

To improve the prediction performance of the surrogate function, \method transfers meta-surrogate functions from 
similar meta-train tasks (see \S \ref{subsec:surrogate_transfer}). 
Fig. \ref{fig:ablation_transfer} shows that this transfer helps find better HPs, demonstrating the added value of meta-learning besides PPE training of \f and surrogate initialization. 

\vspace{-0.15in}

\section{Conclusion}
\label{sec:conclusion}
We introduce (to our knowledge) the {\em first systematic hyperparameter  optimization (HPO) approach for unsupervised outlier detection (OD)}. The proposed \method is a meta-learner, and builds on an extensive pool of \textit{existing} OD benchmark datasets based on which it trains a performance predictor (offline). 
Given a new task without labels (online), 
it capitalizes on the performance predictor 
to enable (originally supervised) sequential model-based optimization for identifying promising HPs iteratively.
Notably, \method stands out from all prior work on \textit{HPO for OD} in being
capable of handling both discrete \textit{and} continuous HPs.
Extensive experiments on three (including both deep and shallow) OD algorithms show its generality, where it significantly outperforms a diverse set of baselines. 
Future work will consider joint 
algorithm selection and continuous hyperparameter optimization for unsupervised outlier detection.
\vspace{-0.1in}

\bibliography{ref}
\vspace{-0.2in}
\bibliographystyle{siam}

\appendix 
\section*{Supplementary Material for \method}
\textit{Details on algorithm design and experiments.}
\setcounter{table}{0}
\setcounter{figure}{0}
\setcounter{algorithm}{0}

\renewcommand{\thetable}{\Alph{section}\arabic{table}}
\renewcommand{\thefigure}{\Alph{section}\arabic{figure}}
\renewcommand{\thealgorithm}{\Alph{section}\arabic{algorithm}}

\section{Details on Supervised SMBO}
\label{appx:SMBO}

Algorithm \ref{algo:smbo} shows the pseudo-code of classical SMBO for HPO. In each iteration, the surrogate function \s predicts the performance and uncertainty of a group of sampled HP settings, where the acquisition function \aq selects the best one to be evaluated next by the objective function $\mathcal{L}(\cdot)$. With the newly evaluated pair of HP settings and the objective value, the surrogate function is updated to be more accurate in iteration. 

\begin{algorithm}[!ht]
	\caption{SMBO for Supervised Hyperparameter Optimization 
	}
	\label{algo:smbo}
	\small{
	\begin{algorithmic}[1]
		\REQUIRE learning algorithm $M$, surrogate function \s, input task $\bD_{\text{test}}=\{\bXs, \mathbf{y}_{\text{test}}\}$,
		objective function $\mathcal{L}(\cdot)$, number of iterations $E$
		\ENSURE selected hyperparameter setting $\vlambda^{*}$ for $\bD_{\text{test}}$
		\algrule
		\STATE Initialize surrogate function $s^{(1)}$
		\FOR{$e=1$ to $E$}
		\STATE $\vlambda^{(e)} := \argmax_{\vlambda \in \vLambda} EI(\bX_{\text{test}}), \vlambda|M,  s^{(e)}$
		\STATE $\mathcal{L}^{(e)} := \text{evaluate} \; \mathcal{L}(M_{\vlambda^{(e)}}, \bXs, \mathbf{y}_{\text{test}})$ \COMMENT{infeasible for \textit{HPO for OD}}
		\STATE Update to $s^{(e+1)}$ with new information $\langle\vlambda^{(e)},\mathcal{L}^{(e)}\rangle$
		\ENDFOR
		\STATE \textbf{Output} $\vlambda^{*} \in \vlambda^{(1)}, \dots, \vlambda^{(e)}$ with the highest evaluated objective function values
	\end{algorithmic}
	}
\end{algorithm}

Clearly, the classical SMBO does not apply to \textit{HPO for OD} directly since the objective function $\mathcal{L}(\cdot)$ cannot be evaluated without ground truth labels (line 4). The proposed \method uses meta-learning to train a regressor \f to predict the performance of an HP on the new dataset without any labels (\S \ref{subsec:meta_train}), and thus enables (originally supervised) SMBO for \textit{HPO for OD} (\S \ref{subsec:online_hpo}).


\section{(Offline) Meta-training Details}
\label{appx:offline}

\subsection{Internal Performance Measures (IPMs).}
\label{appx:ipm}

As described in \S \ref{subsec:f_train}, IPMs are used as part of the input features of the \textit{proxy performance evaluator} \f. In \method, we use three consensus-based IPMs (i.e., MC, SELECT, and HITS) which carry useful and noisy signals in unsupervised OD model selection \cite{ma2021large}. in short, consensus-based IPMs consider the resemblance to the overall consensus of outlier scores as a sign of a better (performance of) model. Thus, a group of models is needed to compute these IPMs for resemblance measure.

In \cite{ma2021large}, they use all models in $\mM$ for building IPMs (i.e., $\mM = \{M_1, \dots, M_m\}$ by pairing detector $M$ with each HP in meta-HP set $\vlambda_{\mathbf{meta}} = \{\vlambda_1, \ldots, \vlambda_m \} \in \vLambda$), leading to high cost in generating outlier scores and then IPMs. To reduce the cost, we instead identify a small subset of representative models $\mM_A \in \mM$ called the \textit{anchor} set (i.e., $|\mM_A|\ll |\mM|$), for calculating IPMs. That is, we generate the IPMs of a model with regard to its consensus to $\mM_A$ rather than $\mM$, for both the meta-train database and the input dataset. The construction of the anchor set can be done by cross-validation in a forward selection way. 

\subsection{Building Proxy Performance Evaluator (\ftrain).}
\label{appx:ftrain}
As outlined in \S \ref{subsec:f_train}, we build \textit{Proxy Performance Evaluator} \f to
map 
\{\texttt{HP setting}, \texttt{meta-features}, \texttt{IPMs}\} of HP $\vlambda_j \in \vlambda_{\mathbf{meta}}$ on the $i$-th meta-train dataset onto its 
ground truth performance, i.e.,
$f(\vlambda_j, \mathbf{m}_i, \bIij) \mapsto \bP_{i,j}$. Given we have $n$ meta-train datasets and the meta-HP set with $	\lvert\vlambda_{\mathbf{meta}}\rvert=m$ HP settings, \f is trained on $mn$ samples by pairing $\vlambda_{\mathbf{meta}}$ with meta-train datasets.


To construct the training samples of \f, we first extract meta-features from each meta-train dataset as
$\bM= \{\mathbf{m}_1, \ldots, \mathbf{m}_n\}=\psi(\{\bX_1, \ldots, \bX_n\})\in \mathbbm{R}^{n \times d}$,
where $\psi(\cdot)$ is the extraction module, and $d$ is the dimension of meta-features.

We also need to extract IPMs of 
each detector $M_j$ with HP setting $\vlambda_j \in \vlambda_{\mathbf{meta}}$ on each meta-train dataset $\mD_i \in \mDt$, where $\bIij := \phi(\mO_{i,j})$ refers to the IPMs using the $j$-th HP setting
the $i$-th meta-train dataset and $\phi$ is the extractor.

Putting these together, we train \f with $(m\cdot n)$ samples. In implementation we use LightGBM \cite{DBLP:conf/nips/KeMFWCMYL17} for \f, while it is flexible to choose any other. 
\vspace{-0.1in}

\subsection{Meta-surrogate Functions (\ffree).}
\label{appx:ffree}
As described in \S \ref{subsec:MSF}, we also train $n$ \textit{independent} regressors with \textit{only HPs as input}, $\bm{\mathcal{T}} = \{t_1, \ldots, t_n\}$. That is, for each meta-train dataset $\mD_i \in \mDt$, we train a regressor $t_i(\cdot)$ that simply maps the $j$-th HP setting $\vlambda_j \in \vlambda_{\mathbf{meta}}$ to its detection performance on the $i$-th meta-train dataset, i.e., $t_i(\vlambda_j) \mapsto \bP_{i,j}$. Thus, $t_i(\cdot)$ only trains on the $m$ HP settings' performance on the $i$-th meta-train dataset. In implementation, we use Gaussian Process (GP) \cite{williams1995gaussian} for \ffree\footref{footnote:gp}, 
and we suggest using the same regressor as the surrogate \s in \S \ref{subsec:online_hpo} for easy knowledge transfer in \S \ref{subsec:surrogate_transfer}.
\vspace{-0.1in}

\section{(Online) Model Selection Details}
\label{appx:online}

\subsection{Sampling Range.}
\label{appx:sampling}

Given the \ftrain, \f, is trained on the meta-HP set $\vlambda_{\mathbf{meta}}$ of the meta-train database, it is more accurate in predicting the HPs from a similar range for the new dataset. Thus, \method samples HPs within the range of $\vlambda_{\mathbf{meta}}$ in SMBO (see \S \ref{subsec:use_smbo}). For instance, given the meta-HP set of iForest shown in Appx. Table \ref{table:rae_iforest}, we sample HPs in range of: (\textit{i}) n\_estimators in $[10,150]$ (\textit{ii}) max\_samples in $[0.1, 0.9]$ and (\textit{iii}) max\_features in $[0.2, 0.8]$ for $\vlambda_{\mathbf{sample}}$. We provide more details on the fast simulation of sampling in Appx. \ref{appx:od_details}.







\section{Additional Exp. Settings and Results}

\subsection{Code and Reproducibility.}
\label{appx:code}

We foster future research by fully releasing the code and the testbed at anonymous repo:  \url{https://tinyurl.com/hpod22}. Upon acceptance, we will release it on GitHub.

\subsection{Datasets.}
\label{appx:datasets}
In Table \ref{table:datasets}, we describe the details of the 39 benchmark datasets used in the experiments---it is composed by 18 datasets from DAMI library \cite{Campos2016evaluation} and 21 datasets from ODDS library \cite{Rayana2016b}.

Note that \method can be extended with more benchmark datasets, and we expect its performance can be further improved.

\begin{table}[!ht]
\caption{Testbed composed by 18 datasets from DAMI library \cite{Campos2016evaluation} and 21 datasets from ODDS library \cite{Rayana2016b}.} 
 	\label{table:datasets} 
\centering
\scalebox{0.75}{
\begin{tabular}{ll|lll}
\toprule
\textbf{Dataset}   & \textbf{Source} & \textbf{\#Samples} & \textbf{\#Dims} & \textbf{\%Outlier} \\
\midrule
1\_ALOI             & DAMI            & 49534             & 27             & 3.04              \\
2\_Annthyroid       & DAMI            & 7129              & 21             & 7.49              \\
3\_Arrhythmia       & DAMI            & 450               & 259            & 45.78             \\
4\_Cardiotocography & DAMI            & 2114              & 21             & 22.04             \\
5\_Glass            & DAMI            & 214               & 7              & 4.21              \\
6\_HeartDisease     & DAMI            & 270               & 13             & 44.44             \\
7\_InternetAds      & DAMI            & 1966              & 1555           & 18.72             \\
8\_PageBlocks       & DAMI            & 5393              & 10             & 9.46              \\
9\_PenDigits        & DAMI            & 9868              & 16             & 0.2               \\
10\_Pima            & DAMI            & 768               & 8              & 34.9              \\
11\_Shuttle         & DAMI            & 1013              & 9              & 1.28              \\
12\_SpamBase        & DAMI            & 4207              & 57             & 39.91             \\
13\_Stamps          & DAMI            & 340               & 9              & 9.12              \\
14\_Waveform        & DAMI            & 3443              & 21             & 2.9               \\
15\_WBC             & DAMI            & 223               & 9              & 4.48              \\
16\_WDBC            & DAMI            & 367               & 30             & 2.72              \\
17\_Wilt            & DAMI            & 4819              & 5              & 5.33              \\
18\_WPBC            & DAMI            & 198               & 33             & 23.74             \\
\midrule
19\_annthyroid      & ODDS            & 7200              & 6              & 7.42              \\
20\_arrhythmia      & ODDS            & 452               & 274            & 14.6              \\
21\_breastw         & ODDS            & 683               & 9              & 34.99             \\
22\_glass           & ODDS            & 214               & 9              & 4.21              \\
23\_ionosphere      & ODDS            & 351               & 33             & 35.9              \\
24\_letter          & ODDS            & 1600              & 32             & 6.25              \\
25\_lympho          & ODDS            & 148               & 18             & 4.05              \\
26\_mammography     & ODDS            & 11183             & 6              & 2.32              \\
27\_mnist           & ODDS            & 7603              & 100            & 9.21              \\
28\_musk            & ODDS            & 3062              & 166            & 3.17              \\
29\_optdigits       & ODDS            & 5216              & 64             & 2.88              \\
30\_pendigits       & ODDS            & 6870              & 16             & 2.27              \\
31\_pima            & ODDS            & 768               & 8              & 34.9              \\
32\_satellite       & ODDS            & 6435              & 36             & 31.64             \\
33\_satimage-2      & ODDS            & 5803              & 36             & 1.22              \\
34\_speech          & ODDS            & 3686              & 400            & 1.65              \\
35\_thyroid         & ODDS            & 3772              & 6              & 2.47              \\
36\_vertebral       & ODDS            & 240               & 6              & 12.5              \\
37\_vowels          & ODDS            & 1456              & 12             & 3.43              \\
38\_wbc             & ODDS            & 378               & 30             & 5.56              \\
39\_wine            & ODDS            & 129               & 13             & 7.75             \\
\bottomrule
\end{tabular}
}

\end{table}

\subsection{OD Algorithms and Hyperparameter Spaces.}
\label{appx:od_details}

We demonstrate the \method effectiveness on three OD algorithms, namely RAE, LOF, and iForest. For RAE, we adapt the author's code with seven key HPs. For LOF and iForest, we use the implementation from Python Outlier Detection (PyOD) library. For fast simulation, we also precompute the outlier scores and IPMs for the inner-HP set (denoted as $\vlambda_{\mathbf{inner}}$), which is within the range of the meta-HP set and serving as additional HPs sampled from continuous HP spaces. In the experiment, \method sets $\vlambda_{\mathbf{sample}}=\vlambda_{\mathbf{meta}} \cup \vlambda_{\mathbf{inner}}$, and uses \s to score all the HPs in $\vlambda_{\mathbf{sample}}$ that are not yet evaluated by \f yet (see \S \ref{subsec:online_hpo}), thus simulating the advantage of sampling from larger ``continuous'' HP spaces. Table \ref{table:all_hps} and the code show details of HP spaces, the meta-HP set, and the inner-HP set.


\begin{table*}[!ht]
\centering
        \caption{Key HPs optimized by \method, and the meta-HP set and the inner-HP set used in this study.}
    \label{table:all_hps}
    \begin{subtable}[t]{\textwidth}
        \centering
        \caption{Key HPs optimized by \method for RAE. Both meta-HP set and inner-HP set include $2\times 3 \times 2 \times 2 \times 2 \times 2 \times 3 $=388 HP settings.}\label{table:rae_hps}

            \scalebox{1}{

        \begin{tabular}{l|ll|ll}
        \toprule
        & \textbf{HP Name} & \textbf{Type}      & \textbf{Meta-HP Set} & \textbf{Inner-HP Set}\\
        \midrule
       1 & \# EncodeLayers   & int (continuous)   & \{2,4\} & \{2,4\}              \\
       2 & Lambda           & float (continuous) & \{5e-5, 5e-3, 5e-1\} & \{1e-4, 1e-3, 1e-1\} \\
       3 &    Learning Rate    & float (continuous) & \{1e-3, 1e-2\}    & \{1e-3, 1e-2\}   \\
       4 &    \# Inner Epochs  & int (continuous)   & \{20, 50\}    & \{30, 40\}       \\
       5 &    \# Outer Epochs  & int (continuous)   & \{20, 50\}    & \{30, 40\}       \\
      6 &    Shrinkage Decay  & int (continuous)   & \{2,4\}        & \{2,4\}      \\
    7 &    Dropout          & float (continuous) & \{0, 0.1, 0.3, 0.5\} 
    & \{0, 0.1, 0.2, 0.4\}    \\
        \bottomrule
        \end{tabular}
    }
    \end{subtable}
    \vspace{0.05in}

    \begin{subtable}[t]{\textwidth}
        \centering
        \caption{Key HPs optimized by \method for LOF. Both meta-HP set and inner-HP set include $40\times 5$=200 HP settings.}\label{table:rae_lof}

            \scalebox{0.835}{

        \begin{tabular}{l|ll|ll}
        \toprule
        & \textbf{HP Name} & \textbf{Type}      & \textbf{Meta-HP Set} & \textbf{Inner-HP Set}\\
        \midrule
       1 & n\_neighbors     & int (continuous)  & \{1,3,5,$\dots$,80\} & \{2,4,6,$\dots$,81\}                                           \\
2 & distance metric           & str (categorical) &  \makecell{\{'chebyshev', 'minkowski', 'cosine', 'euclidean','manhattan'\}} & Same
        \\
        \bottomrule
        \end{tabular}
    }
    \end{subtable}
    \vspace{0.05in}

    \begin{subtable}[t]{\textwidth}
        \centering
        \caption{Key HPs optimized by \method for iForest. Both meta-HP set and inner-HP set include $8\times 9 \times 4$=288 HP settings.}\label{table:rae_iforest}

            \scalebox{0.98}{

        \begin{tabular}{l|ll|ll}
        \toprule
        & \textbf{HP Name} & \textbf{Type}      & \textbf{Meta-HP Set} & \textbf{Inner-HP Set}\\
        \midrule
       1&n\_estimators    & int (continuous)   & \{10,20,30,40,50,75,100,150\} & \{10,20,30,40,50,75,100,150\} \\
2&max\_samples     & float (continuous) & \{0.1, 0.2, $\dots$, 0.9\} & \{0.1, 0.2, $\dots$, 0.9\}            \\
3&max\_features    & float (continuous) & \{0.2, 0.4, 0.6, 0.8\} & \{0.3, 0.5, 0.7, 0.75\}        
        \\
        \bottomrule
        \end{tabular}
    }
    \end{subtable}

\end{table*}

\vfill


\subsection{Baselines.}
\label{appx:baselines}

We provide the details of baselines presented in Table \ref{table:baseline} and \S \ref{subsec:exp_details}, namely simple methods and complex methods.

\noindent \textbf{\textit{Simple methods}}: 
\begin{enumerate}[label={(\arabic*)},leftmargin=0.6cm]
    \item \textbf{Default} always employs the same default/popular HP setting of the underlying OD algorithm (only applicable to the algorithms with recommended HPs).
    \item \textbf{Random} denotes selecting HPs randomly.
    \item \textbf{MinLoss} outputs the HP with the lowest internal loss (only applicable to the algorithms with an internal objective/loss) from a group of random samples HPs.
\end{enumerate}

\noindent \textbf{\textit{Complex methods}}:

\begin{enumerate}[label={(\arabic*)},leftmargin=0.6cm,resume]
    \item \textbf{Hyperensemble (HyperEns or HE)} that averages the outlier scores of randomly sampled HPs \cite{wenzel2020hyperparameter}. Strictly speaking, HE is not an HPO method.
    \item \textbf{Global Best (GB)} selects the best performing HP on meta-train database on average.
    \item \textbf{ISAC}  ~\cite{conf/ecai/KadiogluMST10} first groups meta-train datasets into clusters, and assigns the best performing HP in the meta-HP set to each cluster. During the online HPO phase, it first assigns the new dataset to one of the clusters and uses the group-based HP for the new dataset. 
    \item \textbf{ARGOSMART (AS)} ~\cite{nikolic2013simple} identifies the most similar meta-train dataset of the new task, and outputs the best performing HP on the meta-task for the new task.
    \item \textbf{MetaOD} ~\cite{zhao2021automatic} uses matrix factorization to capture the dataset similarity and HPs' performance similarity, which is the SOTA unsupervised outlier model selection method.
\end{enumerate}

Additionally, we include (9) \methodz, a variant of \method that  directly uses \f to choose from randomly sampled HPs (see \S \ref{subsec:init}).
Note that these unsupervised OD model selection baselines (5-8) are not original for HPO for OD, and could not work with continuous HP spaces. We adapt them for HPO by selecting an HP from the meta-HP set described in \S \ref{subsec:overview}.

\subsection{Full Trace of Case Study.}
\label{appx:case_study}
\begin{table}[!h]
\centering
\caption{Full trace of \method on \texttt{Cardiotocography} dataset. Over iterations (col. 1), \method gradually identifies better HPs (col. 2 \&3), with higher AP (col. 4). The optimal HP on from the meta-HP set is \{'Chebyshev', 79\}, which \method gets closer to the optimal HP during its adaptive search (i.e., finding \{'Chebyshev', 73\} in 30 iterations).} 
	\label{table:case_study_full} 
 \footnotesize
	
\scalebox{0.92}{
\begin{tabular}{l|ll|l}
\toprule
\textbf{\# Iter} & \textbf{Metrics} & \textbf{\# Neighbors} & \textbf{Norm. AP Rank} \\
\midrule
1                & Manhattan         & 23                    & 0.2866          \\
2                & Manhattan         & 23                    & 0.2866          \\
3                & Manhattan         & 23                    & 0.2866          \\
4                & Manhattan         & 23                    & 0.2866          \\
5                & Cosine           & 41                    & 0.327           \\
6                & Cosine           & 42                    & 0.327           \\
7                & Cosine           & 55                    & 0.3438          \\
8                & Cosine           & 55                    & 0.3438          \\
9                & Cosine           & 55                    & 0.3438          \\
10               & Cosine           & 55                    & 0.3438          \\
11               & Cosine           & 55                    & 0.3438          \\
12               & Cosine           & 55                    & 0.3438          \\
13               & Cosine           & 55                    & 0.3438          \\
14               & Chebyshev        & 72                    & 0.3569          \\
15               & Chebyshev        & 72                    & 0.3569          \\
16               & Chebyshev        & 72                    & 0.3569          \\
17               & Chebyshev        & 72                    & 0.3569          \\
18               & Chebyshev        & 72                    & 0.3569          \\
19               & Chebyshev        & 72                    & 0.3569          \\
20               & Chebyshev        & 72                    & 0.3569          \\
21               & Chebyshev        & 72                    & 0.3569          \\
22               & Chebyshev        & 73                    & 0.357           \\
23               & Chebyshev        & 73                    & 0.357           \\
24               & Chebyshev        & 73                    & 0.357           \\
25               & Chebyshev        & 73                    & 0.357           \\
26               & Chebyshev        & 73                    & 0.357           \\
27               & Chebyshev        & 73                    & 0.357           \\
28               & Chebyshev        & 73                    & 0.357           \\
29               & Chebyshev        & 73                    & 0.357           \\
30               & Chebyshev        & 73                    & 0.357           \\
31               & Chebyshev        & 73                    & 0.357           \\
32               & Chebyshev        & 73                    & 0.357           \\
33               & Chebyshev        & 73                    & 0.357           \\
34               & Chebyshev        & 73                    & 0.357           \\
35               & Chebyshev        & 73                    & 0.357           \\
36               & Chebyshev        & 73                    & 0.357           \\
37               & Chebyshev        & 73                    & 0.357           \\
38               & Chebyshev        & 73                    & 0.357           \\
39               & Chebyshev        & 73                    & 0.357           \\
40               & Chebyshev        & 73                    & 0.357           \\
41               & Chebyshev        & 73                    & 0.357           \\
42               & Chebyshev        & 73                    & 0.357           \\
43               & Chebyshev        & 73                    & 0.357           \\
44               & Chebyshev        & 73                    & 0.357           \\
45               & Chebyshev        & 73                    & 0.357           \\
46               & Chebyshev        & 73                    & 0.357           \\
47               & Chebyshev        & 73                    & 0.357           \\
48               & Chebyshev        & 73                    & 0.357           \\
49               & Chebyshev        & 73                    & 0.357           \\
50               & Chebyshev        & 73                    & 0.357           \\
\midrule\midrule
\textbf{Optimal} & Chebyshev        & 79                    & \textbf{0.3609}                  \\
\bottomrule          
\end{tabular}
}

\end{table}

\vfill 
\subsection{Full Performance Results.}

In addition to the avg. rank plot in Fig. \ref{fig:cd_plots}, we provide the full performance of RAE in Table \ref{table:full_rae_results}, as well as the results for LOF and iForest in Table \ref{table:full_lof_results} and \ref{table:full_iforest_results}, respectively.
\label{appx:full_results}
\begin{table*}[!ht]
    \footnotesize

    \scalebox{0.85}{
    \begin{tabular}{l|lllllll|ll}
    \toprule
    \textbf{datasets}   & \textbf{Random} & \textbf{GB}         & \textbf{ISAC}       & \textbf{AS}         & \textbf{HyperEns}   & \textbf{MetaOD}     & \textbf{MinLoss}    & \textbf{HPOD\_0}    & \textbf{HPOD}       \\
    \midrule
    1\_ALOI             & 0.8234 (3)      & 0.6445 (4)          & 0.6445 (4)          & 0.0612 (8)          & 0.6286 (6)          & 0.0612 (8)          & 0.3747 (7)          & 0.9206 (2)          & \textbf{0.974 (1)}  \\
    2\_Annthyroid       & 0.5506 (5)      & 0.0404 (9)          & 0.4336 (6)          & 0.5508 (4)          & 0.2499 (8)          & 0.8659 (2)          & 0.5651 (3)          & 0.3211 (7)          & \textbf{0.9883 (1)} \\
    3\_Arrhythmia       & 0.4688 (9)      & 0.8112 (5)          & 0.7435 (6)          & \textbf{0.9987 (1)} & 0.4836 (8)          & 0.9036 (3)          & 0.7013 (7)          & 0.9451 (2)          & 0.9036 (3)          \\
    4\_Cardiotocography & 0.6519 (5)      & 0.5951 (6)          & 0.819 (3)           & 0.1432 (9)          & 0.6727 (4)          & 0.2786 (8)          & 0.5544 (7)          & 0.949 (2)           & \textbf{0.9688 (1)} \\
    5\_Glass            & 0.6403 (3)      & 0.5781 (6)          & 0.4036 (7)          & 0.1823 (9)          & \textbf{0.7065 (1)} & 0.6094 (5)          & 0.2651 (8)          & 0.6549 (2)          & 0.6354 (4)          \\
    6\_HeartDisease     & 0.5156 (3)      & 0.3958 (6)          & 0.5156 (2)          & 0.056 (8)           & \textbf{0.5662 (1)} & 0.1901 (7)          & 0.4128 (5)          & 0.4487 (4)          & 0.056 (8)           \\
    7\_InternetAds      & 0.6623 (2)      & 0.5156 (7)          & 0.5156 (7)          & 0.5221 (6)          & \textbf{0.6743 (1)} & 0.5404 (5)          & 0.3419 (9)          & 0.6549 (3)          & 0.651 (4)           \\
    8\_PageBlocks       & 0.7584 (5)      & 0.7591 (3)          & 0.7591 (3)          & \textbf{0.9779 (1)} & 0.6078 (7)          & 0.3047 (8)          & 0.7544 (6)          & 0.3047 (8)          & 0.9505 (2)          \\
    9\_PenDigits        & 0.4909 (4)      & 0.0391 (9)          & 0.1289 (7)          & 0.4909 (5)          & \textbf{0.8 (1)}    & 0.4909 (5)          & 0.5247 (3)          & 0.619 (2)           & 0.1289 (7)          \\
    10\_Pima            & 0.4117 (8)      & 0.8698 (3)          & 0.8698 (3)          & 0.9779 (1)          & 0.7792 (5)          & 0.4115 (9)          & 0.6865 (6)          & 0.5247 (7)          & \textbf{0.9779 (1)} \\
    11\_Shuttle         & 0.3195 (2)      & 0.319 (3)           & 0.319 (3)           & 0.319 (3)           & \textbf{0.3849 (1)} & 0.319 (3)           & 0.2464 (7)          & 0.0784 (8)          & 0.056 (9)           \\
    12\_SpamBase        & 0.3208 (6)      & 0.1276 (8)          & 0.1276 (8)          & \textbf{0.9323 (1)} & 0.4618 (4)          & 0.7214 (2)          & 0.5242 (3)          & 0.194 (7)           & 0.3364 (5)          \\
    13\_Stamps          & 0.6727 (5)      & 0.6302 (6)          & 0.1133 (9)          & 0.5924 (7)          & 0.7777 (4)          & \textbf{0.9844 (1)} & 0.5318 (8)          & 0.9026 (3)          & 0.9714 (2)          \\
    14\_Waveform        & 0.687 (8)       & 0.7018 (7)          & 0.8242 (5)          & 0.8932 (2)          & 0.5257 (9)          & 0.8932 (2)          & 0.7977 (6)          & \textbf{0.9513 (1)} & 0.8932 (2)          \\
    15\_WBC             & 0.3403 (4)      & 0.3398 (5)          & \textbf{0.8242 (1)} & 0.1628 (8)          & 0.6966 (2)          & 0.6328 (3)          & 0.0648 (9)          & 0.2628 (6)          & 0.2435 (7)          \\
    16\_WDBC            & 0.5896 (5)      & 0.3542 (7)          & 0.0378 (9)          & 0.9115 (2)          & 0.5849 (6)          & 0.3542 (7)          & 0.7099 (4)          & 0.7219 (3)          & \textbf{0.9935 (1)} \\
    17\_Wilt            & 0.5013 (8)      & 0.5013 (1)          & \textbf{0.5013 (1)} & \textbf{0.5013 (1)} & 0.0026 (9)          & \textbf{0.5013 (1)} & 0.5013 (1)          & \textbf{0.5013 (1)} & \textbf{0.5013 (1)} \\
    18\_WPBC            & 0.5039 (5)      & 0.2943 (8)          & 0.5846 (4)          & 0.737 (2)           & 0.2899 (9)          & 0.6354 (3)          & \textbf{0.7562 (1)} & 0.4445 (7)          & 0.5039 (5)          \\
    19\_annthyroid      & 0.6351 (7)      & 0.3112 (8)          & \textbf{0.9922 (1)} & 0.9701 (2)          & 0.8379 (5)          & 0.8307 (6)          & 0.131 (9)           & 0.8703 (4)          & 0.9312 (3)          \\
    20\_arrhythmia      & 0.4753 (9)      & 0.6432 (7)          & 0.9714 (2)          & 0.9596 (4)          & 0.5143 (8)          & 0.7344 (5)          & 0.6883 (6)          & 0.969 (3)           & \textbf{0.9844 (1)} \\
    21\_breastw         & 0.139 (6)       & 0.0716 (9)          & 0.6719 (2)          & 0.138 (7)           & \textbf{0.9818 (1)} & 0.138 (7)           & 0.2052 (5)          & 0.6719 (2)          & 0.6714 (4)          \\
    22\_glass           & 0.4584 (6)      & 0.6589 (3)          & 0.6693 (2)          & 0.375 (8)           & 0.5475 (4)          & \textbf{0.7865 (1)} & 0.1318 (9)          & 0.3802 (7)          & 0.5312 (5)          \\
    23\_ionosphere      & 0.7039 (9)      & 0.8047 (6)          & 0.8216 (5)          & \textbf{0.9844 (1)} & 0.7143 (8)          & 0.7969 (7)          & 0.8307 (4)          & 0.9096 (3)          & 0.9351 (2)          \\
    24\_letter          & 0.5013 (3)      & 0.0872 (7)          & 0.0182 (8)          & 0.0182 (8)          & 0.5273 (2)          & \textbf{0.918 (1)}  & 0.4492 (4)          & 0.1643 (6)          & 0.2591 (5)          \\
    25\_lympho          & 0.7506 (8)      & 0.7695 (7)          & 0.8802 (5)          & 0.9583 (3)          & 0.6545 (9)          & 0.974 (2)           & 0.7805 (6)          & 0.9177 (4)          & \textbf{1 (1)}      \\
    26\_mammography     & 0.4844 (6)      & 0.3607 (9)          & 0.4844 (7)          & 0.3841 (8)          & 0.5896 (4)          & \textbf{0.9349 (1)} & 0.7328 (3)          & 0.531 (5)           & 0.7865 (2)          \\
    27\_mnist           & 0.6052 (6)      & \textbf{0.9818 (1)} & 0.7786 (3)          & 0.2513 (9)          & 0.5496 (7)          & 0.6406 (5)          & 0.4677 (8)          & 0.7279 (4)          & 0.9401 (2)          \\
    28\_musk            & 0.5584 (8)      & 0.901 (2)           & 0.6289 (6)          & \textbf{0.9857 (1)} & 0.5922 (7)          & 0.7018 (5)          & 0.1414 (9)          & 0.8112 (4)          & 0.8516 (3)          \\
    29\_optdigits       & 0.5286 (5)      & 0.1406 (8)          & 0.5977 (3)          & 0.2122 (7)          & 0.6509 (2)          & 0.5977 (3)          & \textbf{0.794 (1)}  & 0.2721 (6)          & 0.0755 (9)          \\
    30\_pendigits       & 0.626 (3)       & 0.5169 (6)          & 0.5299 (5)          & 0.1445 (8)          & \textbf{0.8119 (1)} & 0.0924 (9)          & 0.318 (7)           & 0.7096 (2)          & 0.613 (4)           \\
    31\_pima            & 0.713 (4)       & 0.7135 (3)          & 0.569 (7)           & 0.569 (7)           & 0.6312 (6)          & 0.569 (7)           & 0.6779 (5)          & 0.9247 (2)          & \textbf{0.9987 (1)} \\
    32\_satellite       & 0.8714 (4)      & 0.8503 (6)          & 0.8685 (5)          & \textbf{0.9909 (1)} & 0.7813 (7)          & 0.2865 (9)          & 0.7471 (8)          & 0.9471 (2)          & 0.9323 (3)          \\
    33\_satimage-2      & 0.9143 (3)      & 0.8333 (6)          & 0.8737 (5)          & \textbf{0.987 (1)}  & 0.7927 (7)          & 0.3997 (9)          & 0.481 (8)           & 0.9063 (4)          & 0.9506 (2)          \\
    34\_speech          & 0.6299 (7)      & 0.9714 (3)          & 0.4154 (9)          & 0.9792 (2)          & 0.4842 (8)          & 0.8177 (4)          & 0.6773 (6)          & 0.699 (5)           & \textbf{0.9909 (1)} \\
    35\_thyroid         & 0.6468 (8)      & 0.9245 (4)          & 0.8633 (6)          & 0.9453 (2)          & \textbf{0.9647 (1)} & 0.7305 (7)          & 0.2867 (9)          & 0.9164 (5)          & 0.9247 (3)          \\
    36\_vertebral       & 0.6974 (8)      & 0.6979 (3)          & 0.6979 (3)          & 0.6979 (3)          & 0.4649 (9)          & 0.6979 (3)          & 0.7487 (2)          & \textbf{0.8133 (1)} & 0.6979 (3)          \\
    37\_vowels          & 0.7506 (7)      & 0.9349 (4)          & 0.4648 (9)          & \textbf{0.9974 (1)} & 0.6987 (8)          & 0.8581 (6)          & 0.869 (5)           & 0.9846 (2)          & 0.9818 (3)          \\
    38\_wbc             & 0.6195 (6)      & \textbf{0.9401 (1)} & 0.8815 (3)          & 0.5443 (7)          & 0.6566 (5)          & 0.5443 (7)          & 0.456 (9)           & 0.9117 (2)          & 0.8727 (4)          \\
    39\_wine            & 0.4818 (4)      & 0.4818 (5)          & 0.4818 (5)          & 0.4818 (5)          & 0.9403 (2)          & \textbf{0.9714 (1)} & 0.5797 (3)          & 0.387 (9)           & 0.4818 (5)          \\
    \midrule
    Average             & 0.5821 (7)      & 0.567 (8)           & 0.5981 (6)          & 0.6047 (5)          & 0.6226 (3)          & \textbf{0.6082 (4)} & 0.5258 (9)          & 0.6622 (2)          & \textbf{0.7216 (1)} \\
    STD                 & 0.1580          & 0.2884              & 0.2630              & 0.3478              & 0.1937              & 0.2647              & 0.2261              & 0.2718              & 0.3103             \\
    \midrule
    Avg. Rank & 5.5641&	5.4103&	4.8462&	4.4359&	5.0513&	4.7949	&5.7949	&4.0256&	\textbf{3.3333}
 \\
    \bottomrule
    \end{tabular}
    }
    \caption{Method evaluation on RAE (normalized AP rank). The most performing method is highlighted in bold. The algo. rank is provided in parenthesis (lower ranks denote better performance). \method achieves the best performance among all baselines.}    \label{table:full_rae_results}

\end{table*}
\begin{table*}[!ht]
    \footnotesize

    \scalebox{0.85}{
    \begin{tabular}{l|lllllll|ll}
    \toprule
    \textbf{datasets}   & \textbf{Random} & \textbf{GB}         & \textbf{ISAC}       & \textbf{AS}         & \textbf{HyperEns}   & \textbf{MetaOD}     & \textbf{Default}    & \textbf{HPOD\_0}    & \textbf{HPOD}       \\
    \midrule
1\_ALOI             & 0.6418 (5)      & 0.6825 (4)          & 0.625 (6)           & \textbf{0.95 (1)}  & 0.005 (9)           & 0.1319 (8)          & 0.905 (2)           & 0.8975 (3)          & \textbf{0.3433 (7)} \\
2\_Annthyroid       & 0.5224 (7)      & \textbf{0.6175 (5)} & 0.565 (6)           & 0.0275 (8)         & 0.8836 (3)          & 0.9236 (2)          & \textbf{0.97 (1)}   & 0.0125 (9)          & \textbf{0.6841 (4)} \\
3\_Arrhythmia       & 0.3134 (8)      & \textbf{0.4125 (5)} & 0.4 (6)             & \textbf{0.98 (1)}  & 0.8557 (3)          & 0.2188 (9)          & 0.335 (7)           & 0.92 (2)            & 0.7761 (4)          \\
4\_Cardiotocography & 0.4776 (4)      & 0.31 (6)            & \textbf{0.745 (3)}  & 0.19 (8)           & \textbf{1 (1)}      & 0.4583 (5)          & 0.165 (9)           & 0.2225 (7)          & \textbf{0.985 (2)}  \\
5\_Glass            & 0.5721 (3)      & 0.3975 (5)          & 0.725 (2)           & 0.02 (9)           & \textbf{0.2647 (7)} & 0.1684 (8)          & \textbf{0.95 (1)}   & 0.3525 (6)          & 0.5498 (4)          \\
6\_HeartDisease     & 0.5473 (6)      & 0.7625 (3)          & 0.4825 (7)          & 0.035 (9)          & \textbf{1 (1)}      & 0.6319 (4)          & 0.32 (8)            & 0.555 (5)           & \textbf{0.8507 (2)} \\
7\_InternetAds      & 0.5174 (5)      & 0.5675 (3)          & \textbf{1 (1)}      & 0.39 (7)           & \textbf{0.4825 (6)} & \textbf{0.9132 (2)} & 0.02 (9)            & 0.2775 (8)          & 0.5572 (4)          \\
8\_PageBlocks       & 0.4129 (6)      & 0.4425 (5)          & \textbf{0.985 (1)}  & \textbf{0.235 (8)} & 0.61 (3)            & 0.5903 (4)          & 0.24 (7)            & 0.155 (9)           & \textbf{0.8507 (2)} \\
9\_PenDigits        & 0.3333 (7)      & 0.3475 (6)          & 0.69 (4)            & \textbf{0.4 (5)}   & \textbf{0.1333 (8)} & 0.8924 (3)          & \textbf{0.9 (1)}    & 0.075 (9)           & \textbf{0.9 (1)}    \\
10\_Pima            & 0.3184 (8)      & 0.535 (6)           & \textbf{0.9975 (1)} & 0.47 (7)           & 0.99 (2)            & 0.783 (4)           & 0.02 (9)            & 0.59 (5)            & \textbf{0.9125 (3)} \\
11\_Shuttle         & 0.5174 (6)      & 0.2725 (8)          & 0.52 (5)            & 0.6925 (3)         & \textbf{0.0229 (9)} & 0.309 (7)           & \textbf{0.895 (1)}  & \textbf{0.7125 (2)} & 0.6925 (3)          \\
12\_SpamBase        & 0.403 (5)       & 0.4075 (4)          & 0.21 (6)            & \textbf{0.17 (8)}  & \textbf{1 (1)}      & 0.1823 (7)          & \textbf{0.155 (9)}  & 0.715 (3)           & 0.8259 (2)          \\
13\_Stamps          & 0.408 (4)       & 0.4225 (3)          & 0.8 (2)             & 0.08 (9)           & 0.408 (4)           & \textbf{0.1406 (8)} & 0.195 (6)           & 0.1625 (7)          & \textbf{0.815 (1)}  \\
14\_Waveform        & 0.7761 (3)      & 0.2025 (6)          & \textbf{0.3425 (5)} & 0.175 (7)          & 0.0338 (9)          & 0.6528 (4)          & 0.85 (2)            & \textbf{0.0575 (8)} & \textbf{0.975 (1)}  \\
15\_WBC             & 0.5522 (5)      & 0.4175 (6)          & \textbf{0.99 (1)}   & 0.01 (9)           & 0.9284 (3)          & \textbf{0.9653 (2)} & 0.195 (7)           & 0.1775 (8)          & 0.765 (4)           \\
16\_WDBC            & 0.1592 (8)      & 0.6575 (3)          & 0.755 (2)           & 0.6575 (3)         & 0.1095 (9)          & \textbf{0.8368 (1)} & 0.235 (6)           & 0.6575 (3)          & \textbf{0.2 (7)}    \\
17\_Wilt            & 0.4378 (4)      & 0.645 (3)           & \textbf{0.185 (7)}  & \textbf{1 (1)}     & 0.005 (9)           & \textbf{0.4045 (5)} & 0.2 (6)             & \textbf{0.995 (2)}  & \textbf{0.17 (8)}   \\
18\_WPBC            & 0.2886 (8)      & 0.695 (4)           & 0.7575 (2)          & 0.54 (6)           & 0.2527 (9)          & 0.3385 (7)          & \textbf{0.7275 (3)} & \textbf{0.95 (1)}   & \textbf{0.615 (5)}  \\
19\_annthyroid      & 0.4428 (8)      & 0.2525 (9)          & \textbf{0.58 (6)}   & \textbf{1 (1)}     & 0.9343 (2)          & 0.7743 (3)          & 0.485 (7)           & \textbf{0.61 (5)}   & 0.7475 (4)          \\
20\_arrhythmia      & 0.4328 (6)      & 0.4425 (5)          & 0.165 (9)           & 0.94 (2)           & 0.7055 (3)          & 0.6927 (4)          & 0.205 (7)           & 0.19 (8)            & \textbf{0.99 (1)}   \\
21\_breastw         & 0.6318 (6)      & 0.0975 (8)          & 0.8 (3)             & 0.7025 (4)         & \textbf{1 (1)}      & 0.0556 (9)          & 0.185 (7)           & 0.8675 (2)          & 0.7025 (4)          \\
22\_glass           & 0.5473 (3)      & 0.5175 (4)          & 0.105 (8)           & 0.325 (7)          & 0.0229 (9)          & \textbf{0.3524 (6)} & \textbf{0.91 (2)}   & \textbf{0.9825 (1)} & 0.4428 (5)          \\
23\_ionosphere      & 0.3781 (5)      & 0.5125 (4)          & \textbf{0.985 (1)}  & \textbf{0.77 (3)}  & 0.205 (7)           & 0.0694 (8)          & 0.055 (9)           & 0.2825 (6)          & \textbf{0.9403 (2)} \\
24\_letter          & 0.5224 (5)      & 0.7425 (3)          & 0.055 (7)           & 0.005 (8)          & 0.005 (9)           & \textbf{0.9722 (1)} & 0.95 (2)            & 0.44 (6)            & 0.54 (4)            \\
25\_lympho          & 0.4428 (6)      & 0.8275 (2)          & 0.03 (9)            & 0.3 (7)            & \textbf{0.9721 (1)} & \textbf{0.8108 (3)} & 0.6375 (5)          & 0.2425 (8)          & \textbf{0.805 (4)}  \\
26\_mammography     & 0.4129 (6)      & 0.4175 (5)          & \textbf{0.9075 (2)} & 0.215 (7)          & \textbf{1 (1)}      & \textbf{0.6563 (4)} & 0.045 (9)           & 0.1825 (8)          & 0.6775 (3)          \\
27\_mnist           & 0.5025 (5)      & \textbf{0.4175 (6)} & 0.24 (8)            & 0.055 (9)          & 0.7522 (2)          & 0.6302 (3)          & 0.505 (4)           & 0.405 (7)           & \textbf{0.79 (1)}   \\
28\_musk            & 0.7662 (2)      & 0.6475 (4)          & \textbf{0.59 (5)}   & \textbf{0.155 (8)} & \textbf{1 (1)}      & 0.066 (9)           & 0.675 (3)           & 0.3575 (7)          & 0.3881 (6)          \\
29\_optdigits       & 0.5572 (7)      & 0.8925 (3)          & 0.3225 (9)          & \textbf{1 (1)}     & 0.6299 (6)          & 0.7465 (5)          & \textbf{0.765 (4)}  & 0.905 (2)           & \textbf{0.45 (8)}   \\
30\_pendigits       & 0.6219 (8)      & 0.6625 (6)          & 0.6275 (7)          & 0.1725 (9)         & \textbf{1 (1)}      & \textbf{1 (1)}      & 0.675 (5)           & 0.7525 (4)          & 0.77 (3)            \\
31\_pima            & 0.4776 (6)      & 0.4225 (7)          & \textbf{0.9525 (1)} & \textbf{0.92 (2)}  & 0.3775 (8)          & 0.5729 (5)          & 0.05 (9)            & 0.8875 (3)          & \textbf{0.86 (4)}   \\
32\_satellite       & 0.408 (4)       & 0.2725 (8)          & 0.36 (7)            & \textbf{0.905 (3)} & \textbf{1 (1)}      & 0.4063 (5)          & \textbf{0.25 (9)}   & 0.365 (6)           & 0.985 (2)           \\
33\_satimage-2      & 0.5771 (6)      & 0.3325 (8)          & 0.7725 (5)          & \textbf{0.98 (1)}  & \textbf{0.855 (4)}  & 0.5226 (7)          & 0.94 (2)            & 0.915 (3)           & 0.005 (9)           \\
34\_speech          & 0.8209 (3)      & 0.8275 (2)          & 0.57 (5)            & 0.3275 (8)         & \textbf{0.8945 (1)} & 0.349 (7)           & 0.025 (9)           & 0.42 (6)            & \textbf{0.6825 (4)} \\
35\_thyroid         & 0.4925 (7)      & \textbf{0.2925 (8)} & 0.925 (3)           & \textbf{1 (1)}     & \textbf{0.999 (2)}  & 0.5556 (6)          & 0.24 (9)            & 0.58 (5)            & 0.91 (4)            \\
36\_vertebral       & 0.3881 (6)      & 0.6975 (4)          & 0.045 (9)           & 0.135 (8)          & 0.2418 (7)          & 0.8715 (2)          & 0.7375 (3)          & \textbf{0.885 (1)}  & \textbf{0.49 (5)}   \\
37\_vowels          & 0.403 (5)       & 0.2875 (7)          & 0.385 (6)           & \textbf{0.63 (3)}  & 0.0259 (9)          & 0.8403 (2)          & 0.055 (8)           & 0.515 (4)           & \textbf{0.9652 (1)} \\
38\_wbc             & 0.3731 (6)      & \textbf{0.5575 (4)} & 0.435 (5)           & 0.19 (9)           & \textbf{0.9771 (1)} & 0.7326 (3)          & \textbf{0.2 (8)}    & 0.32 (7)            & 0.9478 (2)          \\
39\_wine            & 0.3632 (5)      & \textbf{0.5875 (2)} & 0.44 (4)            & 0.055 (8)          & 0.1512 (7)          & \textbf{0.6806 (1)} & 0.21 (6)            & 0.0425 (9)          & 0.5275 (3)          \\
\bottomrule
Average             & 0.4811 (7)      & 0.5001 (6)          & 0.5658 (3)          & 0.4565 (8)         & 0.5829 (2)          & \textbf{0.5615 (4)} & 0.4379 (9)          & 0.5034 (5)          & \textbf{0.6945 (1)} \\
STD                 & 0.1354          & 0.1911              & 0.3005              & 0.3659             & 0.3996              & 0.2903              & 0.3397              & 0.3122              & 0.2457              \\
    \midrule
    Avg. Rank & 5.5641&	4.9744	&4.7692&	5.5897&	4.5897	&4.7179	&5.6667	&5.2564	&\textbf{3.6667}

\\
    \bottomrule
    \end{tabular}
    }
    \caption{Method evaluation on LOF (normalized AP rank). The most performing method is highlighted in bold. The algo. rank is provided in parenthesis (lower ranks denote better performance). \method achieves the best performance among all baselines.}
        \label{table:full_lof_results}

\end{table*}
\begin{table*}[!ht]
    \footnotesize

    \scalebox{0.85}{
    \begin{tabular}{l|lllllll|ll}
    \toprule
    \textbf{datasets}   & \textbf{Random} & \textbf{GB}         & \textbf{ISAC}       & \textbf{AS}         & \textbf{HyperEns}   & \textbf{MetaOD}     & \textbf{Default}    & \textbf{HPOD\_0}    & \textbf{HPOD}       \\
    \midrule
    1\_ALOI             & 0.3979 (3)      & 0.0191 (8)          & 0.5208 (2)          & 0.0868 (7)          & 0.2789 (5)          & 0.1319 (6)          & 0.0035 (9)          & 0.359 (4)           & \textbf{0.7326 (1)} \\
    2\_Annthyroid       & 0.654 (6)       & \textbf{0.9583 (1)} & 0.0764 (9)          & 0.7535 (5)          & 0.5405 (7)          & 0.9236 (3)          & 0.8789 (4)          & 0.5167 (8)          & \textbf{0.9358 (2)} \\
    3\_Arrhythmia       & 0.4879 (6)      & \textbf{0.8924 (1)} & 0.4427 (7)          & \textbf{0.7517 (4)} & 0.8166 (2)          & 0.2188 (8)          & 0.1107 (9)          & 0.6146 (5)          & 0.8125 (3)          \\
    4\_Cardiotocography & 0.481 (6)       & 0.691 (2)           & \textbf{0.9115 (1)} & 0.6319 (3)          & 0.5765 (4)          & 0.4583 (7)          & 0.5433 (5)          & 0.4115 (9)          & \textbf{0.4444 (8)} \\
    5\_Glass            & 0.6021 (3)      & 0.2517 (8)          & 0.4253 (6)          & 0.3993 (7)          & \textbf{0.5606 (4)} & 0.1684 (9)          & \textbf{0.9585 (1)} & 0.5396 (5)          & 0.7795 (2)          \\
    6\_HeartDisease     & 0.526 (8)       & 0.8681 (3)          & 0.9028 (2)          & 0.6997 (5)          & \textbf{0.6042 (7)} & 0.6319 (6)          & 0.09 (9)            & 0.8326 (4)          & \textbf{0.941 (1)}  \\
    7\_InternetAds      & 0.4498 (7)      & 0.8438 (3)          & 0.2465 (8)          & 0.0174 (9)          & \textbf{0.872 (2)}  & \textbf{0.9132 (1)} & 0.4602 (6)          & 0.6438 (5)          & 0.7543 (4)          \\
    8\_PageBlocks       & 0.3183 (7)      & 0.0972 (8)          & 0.6649 (2)          & \textbf{0.3333 (6)} & 0.5066 (5)          & 0.5903 (4)          & 0.0138 (9)          & 0.6167 (3)          & \textbf{0.9688 (1)} \\
    9\_PenDigits        & 0.481 (4)       & 0.026 (9)           & 0.2951 (6)          & \textbf{0.9878 (1)} & \textbf{0.4976 (3)} & 0.8924 (2)          & 0.0623 (8)          & 0.3108 (5)          & 0.1806 (7)          \\
    10\_Pima            & 0.3599 (7)      & 0.0677 (9)          & 0.9583 (2)          & 0.5417 (6)          & 0.6083 (5)          & 0.783 (3)           & 0.1211 (8)          & 0.6215 (4)          & \textbf{0.9861 (1)} \\
    11\_Shuttle         & 0.4844 (5)      & 0.092 (9)           & 0.5035 (4)          & 0.3472 (6)          & \textbf{0.5654 (3)} & 0.309 (7)           & 0.1886 (8)          & \textbf{0.7326 (1)} & 0.6817 (2)          \\
    12\_SpamBase        & 0.5433 (6)      & 0.8958 (4)          & 0.9028 (3)          & \textbf{0.9479 (2)} & 0.5772 (5)          & 0.1823 (9)          & \textbf{1 (1)}      & 0.2563 (7)          & 0.218 (8)           \\
    13\_Stamps          & 0.6298 (5)      & 0.8941 (2)          & 0.434 (8)           & 0.7795 (4)          & 0.6007 (6)          & \textbf{0.1406 (9)} & 0.564 (7)           & 0.8868 (3)          & \textbf{0.9913 (1)} \\
    14\_Waveform        & 0.4844 (6)      & 0.0799 (8)          & \textbf{0.9757 (1)} & 0.3212 (7)          & 0.546 (5)           & 0.6528 (4)          & 0.0138 (9)          & \textbf{0.8441 (3)} & 0.872 (2)           \\
    15\_WBC             & 0.3737 (7)      & 0.1372 (9)          & \textbf{0.7569 (4)} & 0.7917 (2)          & 0.5273 (6)          & \textbf{0.9653 (1)} & 0.1592 (8)          & 0.7563 (5)          & 0.7917 (2)          \\
    16\_WDBC            & 0.526 (8)       & 0.7188 (4)          & 0.6354 (7)          & 0.6389 (6)          & 0.5038 (9)          & 0.8368 (2)          & 0.6903 (5)          & 0.7955 (3)          & \textbf{0.901 (1)}  \\
    17\_Wilt            & 0.5467 (4)      & 0.2882 (8)          & \textbf{0.3264 (6)} & \textbf{0.3264 (6)} & 0.5827 (3)          & \textbf{0.4045 (5)} & 0.1125 (9)          & \textbf{0.8028 (1)} & \textbf{0.7889 (2)} \\
    18\_WPBC            & 0.4983 (3)      & 0.0799 (8)          & 0.2656 (7)          & 0.0313 (9)          & 0.4997 (2)          & 0.3385 (6)          & \textbf{0.4273 (5)} & 0.4399 (4)          & \textbf{0.7014 (1)} \\
    19\_annthyroid      & 0.5952 (8)      & 0.6528 (7)          & \textbf{0.7292 (3)} & 0.6667 (5)          & 0.5661 (9)          & 0.7743 (2)          & 0.6713 (4)          & \textbf{0.8278 (1)} & 0.6574 (6)          \\
    20\_arrhythmia      & 0.4706 (6)      & 0.3958 (7)          & 0.5399 (5)          & 0.0174 (9)          & 0.6879 (3)          & 0.6927 (2)          & 0.2664 (8)          & 0.5816 (4)          & \textbf{0.8564 (1)} \\
    21\_breastw         & 0.4152 (7)      & 0.599 (5)           & 0.7257 (2)          & 0.5104 (6)          & \textbf{0.8844 (1)} & 0.0556 (9)          & 0.6609 (3)          & 0.6003 (4)          & 0.4115 (8)          \\
    22\_glass           & 0.564 (5)       & 0.1684 (8)          & 0.75 (4)            & 0.0729 (9)          & 0.517 (6)           & \textbf{0.3524 (7)} & \textbf{0.9308 (1)} & 0.7594 (3)          & 0.9201 (2)          \\
    23\_ionosphere      & 0.3218 (5)      & 0.0451 (8)          & 0.1875 (6)          & \textbf{0.941 (1)}  & 0.4637 (4)          & 0.0694 (7)          & 0.0035 (9)          & 0.85 (3)            & \textbf{0.941 (1)}  \\
    24\_letter          & 0.4983 (6)      & 0.224 (7)           & 0.1354 (8)          & 0.7778 (4)          & 0.5384 (5)          & \textbf{0.9722 (1)} & 0.0104 (9)          & 0.7812 (3)          & 0.8201 (2)          \\
    25\_lympho          & 0.4152 (5)      & 0.0556 (9)          & 0.2951 (6)          & 0.6233 (4)          & 0.7772 (2)          & \textbf{0.8108 (1)} & 0.6522 (3)          & 0.2455 (7)          & \textbf{0.1597 (8)} \\
    26\_mammography     & 0.5502 (7)      & 0.6458 (4)          & \textbf{0.7361 (1)} & 0.5521 (6)          & 0.5356 (8)          & \textbf{0.6563 (3)} & 0.7076 (2)          & 0.5267 (9)          & 0.5536 (5)          \\
    27\_mnist           & 0.4187 (6)      & \textbf{0.1285 (8)} & 0.8507 (2)          & 0.2917 (7)          & 0.618 (5)           & 0.6302 (4)          & 0.0035 (9)          & 0.7674 (3)          & \textbf{0.9152 (1)} \\
    28\_musk            & 0.2595 (8)      & 0.9253 (2)          & \textbf{0.9253 (1)} & \textbf{0.3368 (7)} & 0.7661 (5)          & 0.066 (9)           & 0.8443 (4)          & 0.6722 (6)          & 0.9239 (3)          \\
    29\_optdigits       & 0.4879 (7)      & 0.6215 (4)          & 0.6597 (3)          & 0.4861 (8)          & 0.6118 (5)          & 0.7465 (2)          & \textbf{0.6003 (6)} & 0.4281 (9)          & \textbf{0.9325 (1)} \\
    30\_pendigits       & 0.526 (5)       & 0.0764 (9)          & 0.9826 (2)          & 0.3611 (7)          & \textbf{0.6893 (4)} & \textbf{1 (1)}      & 0.8166 (3)          & 0.2007 (8)          & 0.3837 (6)          \\
    31\_pima            & 0.3737 (7)      & 0.3802 (6)          & 0.4514 (4)          & \textbf{0.8628 (1)} & 0.609 (2)           & 0.5729 (3)          & 0.1246 (9)          & 0.4038 (5)          & \textbf{0.2535 (8)} \\
    32\_satellite       & 0.4533 (6)      & 0.9861 (2)          & 0.3247 (8)          & \textbf{0.2535 (9)} & 0.51 (5)            & 0.4063 (7)          & \textbf{0.9965 (1)} & 0.5608 (4)          & 0.7274 (3)          \\
    33\_satimage-2      & 0.436 (3)       & 0.3854 (5)          & 0.0069 (9)          & \textbf{0.4132 (4)} & \textbf{0.7848 (1)} & 0.5226 (2)          & 0.0138 (8)          & 0.266 (6)           & 0.191 (7)           \\
    34\_speech          & 0.6817 (3)      & 0.6458 (4)          & 0.5573 (5)          & 0.5573 (5)          & \textbf{0.9052 (1)} & 0.349 (7)           & 0.0069 (9)          & 0.259 (8)           & \textbf{0.7024 (2)} \\
    35\_thyroid         & 0.5294 (7)      & \textbf{0.9549 (1)} & 0.2309 (9)          & 0.8646 (3)          & \textbf{0.4976 (8)} & 0.5556 (6)          & 0.7197 (5)          & 0.7222 (4)          & 0.8819 (2)          \\
    36\_vertebral       & 0.6367 (7)      & 0.8333 (6)          & 0.9549 (3)          & 0.9757 (2)          & 0.3239 (8)          & 0.8715 (5)          & 0.0208 (9)          & \textbf{0.9125 (4)} & \textbf{0.9861 (1)} \\
    37\_vowels          & 0.5606 (4)      & 0.2639 (8)          & 0.3038 (7)          & \textbf{0.8785 (1)} & 0.6692 (3)          & 0.8403 (2)          & 0.0208 (9)          & 0.4854 (6)          & 0.5382 (5)          \\
    38\_wbc             & 0.4844 (6)      & \textbf{0.1389 (7)} & 0.559 (5)           & 0.1007 (8)          & 0.5599 (4)          & 0.7326 (3)          & \textbf{0.9308 (1)} & 0.8899 (2)          & 0.0313 (9)          \\
    39\_wine            & 0.5917 (4)      & \textbf{0.9063 (1)} & 0.059 (8)           & 0.0104 (9)          & 0.501 (6)           & \textbf{0.6806 (2)} & 0.5709 (5)          & 0.6031 (3)          & 0.3875 (7)          \\
    \midrule
    Average             & 0.4901 (7)      & 0.4598 (8)          & 0.5438 (5)          & 0.5113 (6)          & 0.5969 (3)          & \textbf{0.5615 (4)} & 0.4095 (9)          & 0.5981 (2)          & \textbf{0.6835 (1)} \\
    STD                 & 0.0960          & 0.3493              & 0.2904              & 0.3025              & 0.1364              & 0.2903              & 0.3613              & 0.2087              & 0.2790             \\
    \midrule
    Avg. Rank & 5.7179 &	5.6923&	4.7692&	5.3846&	4.5641	&4.5385	&6.0769&	4.6410&	\textbf{3.5128}
\\
    \bottomrule
    \end{tabular}
    }
    \caption{Method evaluation on iForest (normalized AP rank). The most performing method is highlighted in bold. The algo. rank is provided in parenthesis (lower ranks denote better performance). \method achieves the best performance among all.}
        \label{table:full_iforest_results}

\end{table*}

\label{sec:appendix}

\end{document}